\PassOptionsToPackage{unicode}{hyperref}
\PassOptionsToPackage{hyphens}{url}
\PassOptionsToPackage{dvipsnames,svgnames,x11names}{xcolor}
\documentclass[
]{article}
\usepackage{xcolor}
\usepackage{amsmath,amssymb}
\usepackage{iftex}
\ifPDFTeX
  \usepackage[T1]{fontenc}
  \usepackage[utf8]{inputenc}
  \usepackage{textcomp} 
\else 
  \usepackage{unicode-math} 
  \defaultfontfeatures{Scale=MatchLowercase}
  \defaultfontfeatures[\rmfamily]{Ligatures=TeX,Scale=1}
\fi
\usepackage{lmodern}
\ifPDFTeX\else
\fi
\IfFileExists{upquote.sty}{\usepackage{upquote}}{}
\IfFileExists{microtype.sty}{
  \usepackage[]{microtype}
  \UseMicrotypeSet[protrusion]{basicmath} 
}{}
\makeatletter
\@ifundefined{KOMAClassName}{
  \IfFileExists{parskip.sty}{%
    \usepackage{parskip}
  }{
    \setlength{\parindent}{0pt}
    \setlength{\parskip}{6pt plus 2pt minus 1pt}}
}{
  \KOMAoptions{parskip=half}}
\makeatother
\makeatletter
\ifx\paragraph\undefined\else
  \let\oldparagraph\paragraph
  \renewcommand{\paragraph}{
    \@ifstar
      \xxxParagraphStar
      \xxxParagraphNoStar
  }
  \newcommand{\xxxParagraphStar}[1]{\oldparagraph*{#1}\mbox{}}
  \newcommand{\xxxParagraphNoStar}[1]{\oldparagraph{#1}\mbox{}}
\fi
\ifx\subparagraph\undefined\else
  \let\oldsubparagraph\subparagraph
  \renewcommand{\subparagraph}{
    \@ifstar
      \xxxSubParagraphStar
      \xxxSubParagraphNoStar
  }
  \newcommand{\xxxSubParagraphStar}[1]{\oldsubparagraph*{#1}\mbox{}}
  \newcommand{\xxxSubParagraphNoStar}[1]{\oldsubparagraph{#1}\mbox{}}
\fi
\makeatother

\usepackage{color}
\usepackage{fancyvrb}

\DefineVerbatimEnvironment{Highlighting}{Verbatim}{commandchars=\\\{\}}
\usepackage{framed}
\definecolor{shadecolor}{RGB}{241,243,245}
\newenvironment{Shaded}{\begin{snugshade}}{\end{snugshade}}

\newcommand{\AttributeTok}[1]{\textcolor[rgb]{0.40,0.45,0.13}{#1}}

\newcommand{\BuiltInTok}[1]{\textcolor[rgb]{0.00,0.23,0.31}{#1}}

\newcommand{\CommentTok}[1]{\textcolor[rgb]{0.37,0.37,0.37}{#1}}

\newcommand{\ControlFlowTok}[1]{\textcolor[rgb]{0.00,0.23,0.31}{\textbf{#1}}}

\newcommand{\DecValTok}[1]{\textcolor[rgb]{0.68,0.00,0.00}{#1}}

\newcommand{\FunctionTok}[1]{\textcolor[rgb]{0.28,0.35,0.67}{#1}}
\newcommand{\ImportTok}[1]{\textcolor[rgb]{0.00,0.46,0.62}{#1}}

\newcommand{\KeywordTok}[1]{\textcolor[rgb]{0.00,0.23,0.31}{\textbf{#1}}}
\newcommand{\NormalTok}[1]{\textcolor[rgb]{0.00,0.23,0.31}{#1}}
\newcommand{\OperatorTok}[1]{\textcolor[rgb]{0.37,0.37,0.37}{#1}}
\newcommand{\OtherTok}[1]{\textcolor[rgb]{0.00,0.23,0.31}{#1}}

\newcommand{\SpecialCharTok}[1]{\textcolor[rgb]{0.37,0.37,0.37}{#1}}
\newcommand{\SpecialStringTok}[1]{\textcolor[rgb]{0.13,0.47,0.30}{#1}}
\newcommand{\StringTok}[1]{\textcolor[rgb]{0.13,0.47,0.30}{#1}}
\newcommand{\VariableTok}[1]{\textcolor[rgb]{0.07,0.07,0.07}{#1}}

\usepackage{longtable,booktabs,array}
\usepackage{calc} 
\usepackage{etoolbox}
\makeatletter
\patchcmd\longtable{\par}{\if@noskipsec\mbox{}\fi\par}{}{}
\makeatother
\IfFileExists{footnotehyper.sty}{\usepackage{footnotehyper}}{\usepackage{footnote}}
\makesavenoteenv{longtable}
\usepackage{graphicx}
\makeatletter
\newsavebox\pandoc@box
\newcommand*\pandocbounded[1]{
  \sbox\pandoc@box{#1}%
  \Gscale@div\@tempa{\textheight}{\dimexpr\ht\pandoc@box+\dp\pandoc@box\relax}%
  \Gscale@div\@tempb{\linewidth}{\wd\pandoc@box}%
  \ifdim\@tempb\p@<\@tempa\p@\let\@tempa\@tempb\fi
  \ifdim\@tempa\p@<\p@\scalebox{\@tempa}{\usebox\pandoc@box}%
  \else\usebox{\pandoc@box}%
  \fi%
}
\def\fps@figure{htbp}
\makeatother

\NewDocumentCommand\citeproctext{}{}
\NewDocumentCommand\citeproc{mm}{%
  \begingroup\def\citeproctext{#2}\cite{#1}\endgroup}
\makeatletter
 \let\@cite@ofmt\@firstofone
 \def\@biblabel#1{}
 \def\@cite#1#2{{#1\if@tempswa , #2\fi}}
\makeatother
\newlength{\cslhangindent}
\newlength{\csllabelwidth}
\newenvironment{CSLReferences}[2] 
 {\begin{list}{}{%
  \setlength{\itemindent}{0pt}
  \setlength{\leftmargin}{0pt}
  \setlength{\parsep}{0pt}
  \ifodd #1
   \setlength{\leftmargin}{\cslhangindent}
   \setlength{\itemindent}{-1\cslhangindent}
  \fi
  \setlength{\itemsep}{#2\baselineskip}}}
 {\end{list}}
\usepackage{calc}

\providecommand{\tightlist}{%
  \setlength{\itemsep}{0pt}\setlength{\parskip}{0pt}}

\usepackage{fontspec}
\usepackage{microtype}  

\usepackage{booktabs}
\usepackage{longtable}
\usepackage{array}
\usepackage{tabularx}

\usepackage{graphicx}
\usepackage{float}

\usepackage{etoolbox}

\AtBeginEnvironment{longtable}{\small}

\usepackage[colorlinks=true, linkcolor=black, citecolor=black, urlcolor=blue!60!black]{hyperref}

\makeatletter
\def\ps@fancy{%
  \def\@oddhead{\small\textit{A Grammar of Machine Learning Workflows}\hfill\textit{Roth, 2026}}%
  \def\@evenhead{\small\textit{A Grammar of Machine Learning Workflows}\hfill\textit{Roth, 2026}}%
  \def\@oddfoot{\hfill\thepage\hfill}%
  \def\@evenfoot{\hfill\thepage\hfill}%
}
\makeatother

\newlength{\refhang}
\makeatletter
\@ifpackageloaded{caption}{}{\usepackage{caption}}
\AtBeginDocument{%
\ifdefined\contentsname
  \renewcommand*\contentsname{Table of contents}
\else
  \newcommand\contentsname{Table of contents}
\fi
\ifdefined\listfigurename
  \renewcommand*\listfigurename{List of Figures}
\else
  \newcommand\listfigurename{List of Figures}
\fi
\ifdefined\listtablename
  \renewcommand*\listtablename{List of Tables}
\else
  \newcommand\listtablename{List of Tables}
\fi
\ifdefined\figurename
  \renewcommand*\figurename{Figure}
\else
  \newcommand\figurename{Figure}
\fi
\ifdefined\tablename
  \renewcommand*\tablename{Table}
\else
  \newcommand\tablename{Table}
\fi
}
\@ifpackageloaded{float}{}{\usepackage{float}}
\floatstyle{ruled}
\@ifundefined{c@chapter}{\newfloat{codelisting}{h}{lop}}{\newfloat{codelisting}{h}{lop}[chapter]}
\floatname{codelisting}{Listing}

\makeatother
\makeatletter
\@ifpackageloaded{caption}{}{\usepackage{caption}}
\@ifpackageloaded{subcaption}{}{\usepackage{subcaption}}
\makeatother
\usepackage{bookmark}
\IfFileExists{xurl.sty}{\usepackage{xurl}}{} 
\hypersetup{
  pdftitle={A Grammar of Machine Learning Workflows },
  pdfauthor={Simon Roth},
  pdfkeywords={machine learning, data leakage, grammar, type
system, cross-validation, reproducibility},
  colorlinks=true,
  linkcolor={blue},
  filecolor={Maroon},
  citecolor={Blue},
  urlcolor={Blue},
  pdfcreator={LaTeX via pandoc}}

\title{A Grammar of Machine Learning Workflows\\
\vspace{0.4em}\large\textnormal{Rejecting Data Leakage at Call Time}}
\author{Simon Roth}
\date{May 29, 2026}
\begin{document}
\maketitle
\begin{abstract}
Data leakage has been identified in 648 published papers (as of
mid-2024) across 30 scientific fields
(\citeproc{ref-kapoor2025living}{Kapoor and Narayanan 2025}). The
knowledge to prevent it has existed for over a decade; the problem
persists because the tools do not enforce what the textbooks teach. No
existing ML framework distinguishes formative evaluation (iterative
model development) from summative assessment (terminal performance
judgment) as a type-level distinction, and none rejects the violation at
call time.

This paper presents a grammar (eight typed primitives connected by a
directed acyclic graph with four hard constraints) that makes the most
damaging leakage types structurally unrepresentable within the grammar's
scope. The guarantee covers workflows that stay inside the typed
primitives in a single session on a single \texttt{split}-produced
Partition; manual feature mutation before \texttt{split} and
cross-session re-splits remain outside the guard architecture (see
§Enforcement Gaps for the falsifying workflow). Whether the grammar
reduces leakage errors in human-written code in practice is an empirical
question deferred to a between-subjects study. The core mechanism is a
terminal assessment gate: the first call-time-enforced evaluate/assess
boundary documented in the peer-reviewed ML methodology literature (to
my knowledge, as of May 2026), backed by a specification precise enough
for independent reimplementation. Boundary-aware splitting addresses the
structural layer; the type system catches implementation errors (from
human practitioners, LLM code generators, and AutoML systems alike). A
companion study across 2,047 datasets
(\citeproc{ref-roth2026landscape}{Roth 2026}) grounds the constraints in
measured effect sizes. Three directional predictions recorded before
observing results: two confirmed, one falsified. Two reference
implementations (Python, R) are available.
\end{abstract}

\newpage

\section{Introduction}\label{introduction}

Across 30 scientific fields, 648 published papers have been identified
as containing data leakage errors
(\citeproc{ref-kapoor2025living}{Kapoor and Narayanan 2025}). The
knowledge to prevent the most common of these errors has existed for
over a decade (\citeproc{ref-kaufman2012leakage}{Kaufman et al. 2012};
\citeproc{ref-cawley2010overfitting}{Cawley and Talbot 2010}). The gap
is not knowledge. It is enforcement. The diagnosis aligns with Sculley
et al. (\citeproc{ref-sculley2015hidden}{2015})'s hidden-technical-debt
framing of ML systems: pipeline failures are structural, system-level
debt that conventional tooling and documentation cannot see; closing
them requires changes at the system boundary, not at the practitioner's
discretion.

Documentation does not close structural failures. Best-practice guides
did not prevent a substantial fraction of neuroimaging studies from
leaking (\citeproc{ref-rosenblatt2024data}{Rosenblatt et al. 2024}).
Reporting checklists such as REFORMS and PROBAST-AI did not prevent
leaked results from propagating into field-level meta-analytic
conclusions (\citeproc{ref-vandemortel2025leakage}{van de Mortel and van
Wingen 2025}). The fairness-bias\footnote{\emph{Bias} in this paper
  carries three distinct meanings: \textbf{statistical estimator bias}:
  the systematic offset between an estimator and the parameter it
  targets (e.g., Cawley \& Talbot's ``selection bias'' from
  hyperparameter tuning, nested CV bias, the inflation from picking the
  best of \(K\) noisy evaluations); \textbf{structural bias}:
  effect-estimate distortion from a misspecified sampling process,
  typically under non-iid data (temporal, group, or spatial structure);
  and \textbf{fairness bias}: outcome disparities across protected
  groups (\citeproc{ref-roth2022biased}{Roth 2022}). Leakage is the
  methodological failure that produces statistical and structural bias
  at the workflow level; fairness bias is a downstream concern with its
  own measurement framework, addressed separately.} literature is
scope-adjacent: leakage and fairness bias have distinct root causes
(leakage is a pipeline-isolation failure; fairness bias arises from
label noise, group representation, or distribution skew), but they share
a \emph{measurement dependence}: a model whose training process is
contaminated by test data cannot be trusted to measure fairness either,
because the contaminated measurement is what any downstream fairness
analysis builds on. This paper addresses leakage specifically; fairness
bias is a distinct problem addressed elsewhere
(\citeproc{ref-roth2022biased}{Roth 2022}).

The field responded with documentation: best-practice guides, checklists
(\citeproc{ref-kapoor2024reforms}{Kapoor et al. 2024}), linting tools.
Yang et al. (\citeproc{ref-yang2022leakage}{2022})'s static analysis
catches three leakage types (Overlap, Multi-test, and Preprocessing) at
92.9\% accuracy. Building on Yang's methodology, the LeakageDetector
PyCharm plugin packages the analysis into an IDE
(\citeproc{ref-alomar2025leakagedetector}{AlOmar et al. 2025}), and a
2.0 release extends detection to Jupyter pipelines with LLM-driven
corrections (\citeproc{ref-truong2025leakagedetector2}{Truong et al.
2025}). Drobnjaković, Subotić, and Urban
(\citeproc{ref-drobnjakovic2024abstract}{2024})'s abstract
interpretation tracks partition labels through notebook code at 93\%
precision, the closest existing work to a formal treatment of pipeline
correctness. Both detect leakage. Neither prevents it. Detection finds
the error after the fact; prevention makes it inexpressible. The
question the literature has not answered is whether leakage can be made
structurally inexpressible, rejected by the composition rules of the
workflow itself, not flagged afterward by a separate tool.

\textbf{The code generation gap.} Large language models now generate ML
pipelines faster than any human can review them. Ask one to build a
classifier and it will evaluate on the test set, because that is what
most training examples do. The leaky path is the short path, and the
short path is what language models learn. The present paper addresses
execution-time leakage: what an LLM-generated pipeline does when run;
pretraining-time test-set contamination (where benchmark instances leak
into the model's training corpus, e.g., Sainz et al.
(\citeproc{ref-sainz2023nlpevaluation}{2023}); Deng et al.
(\citeproc{ref-deng2023benchmarkprobing}{2023})) is a separate problem
the grammar's session-scoped guards structurally cannot see. When
pipelines are generated at machine speed, the structural guarantee at
execution time is the only guarantee that remains for the cases the
grammar covers; pretraining contamination requires a different
intervention.

Structural rejection has worked before. Codd
(\citeproc{ref-codd1970relational}{1970}) defined a relational algebra
and a rejection criterion: queries that violate the relational model are
invalid, not merely inadvisable. The five defined operations are
permutation, projection, join, composition, and restriction; Codd's own
adequate-set summary (§2.1) is \emph{projection, natural join, tie, and
restriction}; permutation is irrelevant in the noninferential case and
natural composition is derivable from natural join followed by
projection. SQL implemented this. Wickham
(\citeproc{ref-wickham2010layered}{2010}) did the same for statistical
graphics: ggplot2 exposes a curated subset of Wilkinson
(\citeproc{ref-wilkinson1999grammar}{1999})'s primitives, and incomplete
specifications fail at construction. Both grammars survived because they
had executable implementations. The primitives of the supervised ML
lifecycle are already standardized (sklearn established
fit/predict/transform; tidymodels added per-fold preprocessing). What is
missing is the Wickham move: a curated selection with a rejection
criterion that makes ``this workflow is structurally invalid'' a type
error rather than an opinion. When pipelines are generated at machine
speed (by AutoML systems, LLM code generators, and automated retraining
schedules), structural enforcement is the only guarantee that scales
regardless of who or what writes the code.

\textbf{The Wickham gap.} Wickham
(\citeproc{ref-wickham2010layered}{2010}) presented ggplot2 as a layered
grammar that ``builds on'' but ``differs from Wilkinson's in its
arrangement of the components.'' By choosing which primitives to make
primary and which to omit, Wickham shifted which errors were easy to
make: incomplete specifications fail at construction, and correct
compositions become the default path. The analogous selection has not
been made for the ML lifecycle. Kuhn and Silge
(\citeproc{ref-kuhn2022tidy}{2022}) came closest: recipes inside
tidymodels workflows enforce per-fold preprocessing structurally, and
\texttt{last\_fit()} encourages a one-shot terminal evaluation. The
tidymodels framework got most of the way there. But a user can still fit
a recipe outside a workflow, evaluate on training data, or reuse the
test set. The structure discourages mistakes but does not eliminate
them. There is no type that distinguishes ``this data has been split
correctly'' from ``this data has not.'' The primitives are typed at the
implementation level (recipe, parsnip model, workflow), not at the
correctness level (Partition, PreparedData, Model).

This paper makes that selection for supervised learning, addressing the
question with a \emph{grammar} in the Wilkinson--Wickham sense: a
compositional system with typed primitives and a rejection criterion for
a domain, not a generative rule system in Chomsky's sense (the
distinction is elaborated in Section~\ref{sec-design-properties}). The
underlying mechanism is analogous to information flow control
(\citeproc{ref-myers1999jflow}{Myers 1999}): train, valid, and test
function as security levels, and the guards enforce that test-labeled
data cannot flow into training operations at primitive boundaries
(Section~\ref{sec-design-properties}). Scope is tabular supervised
learning on the four leakage classes identified in the companion study
(\citeproc{ref-roth2026landscape}{Roth 2026}). Temporal leakage is
addressed by boundary-aware primitives (\texttt{split\_temporal},
\texttt{cv\_temporal}); feature engineering is scoped per fold by
\texttt{prepare} after \texttt{split}, and \texttt{fit} rejects
unregistered data that bypassed \texttt{split} entirely; distribution
mismatch across groups is addressed by \texttt{split\_group}. What the
grammar cannot see is contamination baked into the raw data before any
primitive touches it: a feature that encodes the target under a
different name, or a generating process that shifted between collection
and deployment.

The specification (Appendix A) is the paper's core artifact. It defines
the type system, primitive operations, guards, and validity conditions
precisely enough for independent reimplementation. The conformance
conditions are stated in structured English; what it lacks is inference
rules and a soundness proof, but it is complete enough that conformance
is testable from the API surface. Read the Appendix as the
specification; read the rest as the argument for why it should exist.

\section{A Motivating Example}\label{a-motivating-example}

A student who grades their own homework learns nothing new. This code
grades its own homework:

\begin{Shaded}
\begin{Highlighting}[]
\ImportTok{from}\NormalTok{ sklearn.model\_selection }\ImportTok{import}\NormalTok{ train\_test\_split}
\ImportTok{from}\NormalTok{ sklearn.linear\_model }\ImportTok{import}\NormalTok{ LogisticRegression}
\ImportTok{from}\NormalTok{ sklearn.ensemble }\ImportTok{import}\NormalTok{ RandomForestClassifier}
\ImportTok{from}\NormalTok{ sklearn.metrics }\ImportTok{import}\NormalTok{ roc\_auc\_score}

\NormalTok{X\_train, X\_test, y\_train, y\_test }\OperatorTok{=}\NormalTok{ train\_test\_split(X, y)}

\ControlFlowTok{for}\NormalTok{ Model }\KeywordTok{in}\NormalTok{ [LogisticRegression, RandomForestClassifier]:}
\NormalTok{    model }\OperatorTok{=}\NormalTok{ Model().fit(X\_train, y\_train)}
\NormalTok{    p }\OperatorTok{=}\NormalTok{ model.predict\_proba(X\_test)[:, }\DecValTok{1}\NormalTok{]}
    \BuiltInTok{print}\NormalTok{(}\SpecialStringTok{f"}\SpecialCharTok{\{}\NormalTok{Model}\SpecialCharTok{.}\VariableTok{\_\_name\_\_}\SpecialCharTok{\}}\SpecialStringTok{: }\SpecialCharTok{\{}\NormalTok{roc\_auc\_score(y\_test, p)}\SpecialCharTok{:.3f\}}\SpecialStringTok{"}\NormalTok{)}
\end{Highlighting}
\end{Shaded}

Each model's test-set score is computed and compared. The researcher
picks the model with the higher test AUC and reports it. This is
methodologically incorrect: the test set has been used for model
selection, not just final assessment. The measured inflation grows
logarithmically with \(K\), reaching nearly one standardized effect-size
unit at \(K = 19\) in the companion study
(\citeproc{ref-roth2026landscape}{Roth 2026}). No sklearn
(\citeproc{ref-pedregosa2011scikit}{Pedregosa et al. 2011};
\citeproc{ref-buitinck2013sklearn}{Buitinck et al. 2013}) mechanism
prevents this: the API returns plain NumPy arrays; nothing stops the
user from scoring on \texttt{X\_test} repeatedly and choosing the best
result. The textbook rule (``use the test set only once'') is
documentation, not enforcement.

The grammar version:

\begin{Shaded}
\begin{Highlighting}[]
\ImportTok{import}\NormalTok{ ml}

\NormalTok{s }\OperatorTok{=}\NormalTok{ ml.split(df, target}\OperatorTok{=}\StringTok{"y"}\NormalTok{, seed}\OperatorTok{=}\DecValTok{42}\NormalTok{)}

\ControlFlowTok{for}\NormalTok{ algo }\KeywordTok{in}\NormalTok{ [}\StringTok{"logistic"}\NormalTok{, }\StringTok{"rf"}\NormalTok{]:}
\NormalTok{    model }\OperatorTok{=}\NormalTok{ ml.fit(s.train, }\StringTok{"y"}\NormalTok{, algorithm}\OperatorTok{=}\NormalTok{algo, seed}\OperatorTok{=}\DecValTok{42}\NormalTok{)}
\NormalTok{    metrics }\OperatorTok{=}\NormalTok{ ml.evaluate(model, s.valid)}
    \BuiltInTok{print}\NormalTok{(}\SpecialStringTok{f"}\SpecialCharTok{\{}\NormalTok{algo}\SpecialCharTok{\}}\SpecialStringTok{: }\SpecialCharTok{\{}\NormalTok{metrics[}\StringTok{\textquotesingle{}roc\_auc\textquotesingle{}}\NormalTok{]}\SpecialCharTok{:.3f\}}\SpecialStringTok{"}\NormalTok{)}

\CommentTok{\# pick winner, fit on full dev, assess once}
\NormalTok{best }\OperatorTok{=}\NormalTok{ ml.fit(s.dev, }\StringTok{"y"}\NormalTok{, algorithm}\OperatorTok{=}\StringTok{"rf"}\NormalTok{, seed}\OperatorTok{=}\DecValTok{42}\NormalTok{)}
\NormalTok{final }\OperatorTok{=}\NormalTok{ ml.assess(best, test}\OperatorTok{=}\NormalTok{s.test)}
\end{Highlighting}
\end{Shaded}

Each line produces a typed output that constrains what the next line can
accept.

\texttt{split} creates a Partition (train, valid, test, and dev) with
provenance tracked by content hash. \texttt{fit} accepts only data
registered by \texttt{split}; anything else is rejected before training
begins. \texttt{evaluate} checks the model against validation data and
returns Metrics: repeatable, cheap, safe for iteration. This is the
\emph{iterate zone}: the practitioner re-fits and evaluates as many
times as needed.

Only the final model proceeds to \texttt{assess}. It is fit on
\texttt{s.dev} (train + valid combined) to use all available non-test
data. \texttt{assess} returns Evidence: terminal, sealed, irreversible.
A second call on the same holdout fails the guard. The methodologically
incorrect pattern in the sklearn version is now structurally rejected:
the first \texttt{assess} marks the holdout as spent in the provenance
registry, and no amount of cloning, serializing, or restarting recovers
it within the session.

The canonical sklearn leakage example, normalizing before splitting
(Class I), is structurally visible but empirically negligible
(\(d_z \approx 0\)). The grammar prevents it too (unregistered data is
rejected by \texttt{fit}), but the core contribution addresses the
patterns a \texttt{Pipeline} does not: repeated test-set assessment
(\(d_z\) = 0.929, \(\Delta\)AUC = +0.040) and training on evaluation
data (\(d_z\) = 0.372--1.112, \(\Delta\)AUC = +0.004--0.024).

\section{The Grammar}\label{the-grammar}

\subsection{Kernel primitives}\label{kernel-primitives}

Eight operations. That is the entire grammar. Everything a supervised
learning workflow does (splitting, cross-validating, preprocessing,
fitting, predicting, evaluating, explaining, assessing) reduces to these
eight primitives and the types that connect them:

\begin{longtable}[]{@{}
  >{\raggedright\arraybackslash}p{(\linewidth - 6\tabcolsep) * \real{0.0500}}
  >{\raggedright\arraybackslash}p{(\linewidth - 6\tabcolsep) * \real{0.1200}}
  >{\raggedright\arraybackslash}p{(\linewidth - 6\tabcolsep) * \real{0.3500}}
  >{\raggedright\arraybackslash}p{(\linewidth - 6\tabcolsep) * \real{0.4800}}@{}}
\toprule\noalign{}
\begin{minipage}[b]{\linewidth}\raggedright
\#
\end{minipage} & \begin{minipage}[b]{\linewidth}\raggedright
Primitive
\end{minipage} & \begin{minipage}[b]{\linewidth}\raggedright
Type signature
\end{minipage} & \begin{minipage}[b]{\linewidth}\raggedright
What it does
\end{minipage} \\
\midrule\noalign{}
\endhead
\bottomrule\noalign{}
\endlastfoot
1 & \textbf{split} & DataFrame \(\to\) Partition & Create
train/valid/test splits with provenance \\
2 & \textbf{cv} & DataFrame\(^g\) \(\to\) CVResult & Create \(k\)-fold
rotation from training data \\
3 & \textbf{prepare} & DataFrame\(^g\) \(\to\) PreparedData & Normalize,
encode, impute features \\
4 & \textbf{fit} & (DataFrame\(^g\) \(\vert\) CVResult \(\vert\)
PreparedData) \(\times\) target \(\to\) Model & Train a model \\
5 & \textbf{predict} & Model \(\times\) DataFrame \(\to\) Predictions &
Apply model to new data \\
6 & \textbf{evaluate} & Model \(\times\) DataFrame\(^g\) \(\to\) Metrics
& Measure on validation data (repeatable) \\
7 & \textbf{explain} & Model {[}\(\times\) DataFrame{]} \(\to\)
Explanation & Feature importances, partial dependence \\
8 & \textbf{assess} & Model \(\times\) DataFrame\(^g\) \(\to\) Evidence
& Measure on test data (terminal, once) \\
\end{longtable}

\(^g\) Guarded: requires a DataFrame with split provenance (registered
by \texttt{split}).

\texttt{assess} and \texttt{explain} are terminal: their output types
(Evidence, Explanation) feed no further primitive. \texttt{cv}
transforms a training or dev DataFrame into a CVResult, a rotation
schedule of \(k\) train/validation index pairs, separating boundary
creation (\texttt{split}) from iteration (\texttt{cv}). Cross-validation
is a non-test operation: \texttt{cv(s.train)} is the common case;
\texttt{cv(s.dev)} is valid when the practitioner wants to
cross-validate over the full development set. The test partition is
never valid input. In v1.0, cross-validation was a variant inside
Partition; extracting \texttt{cv} into its own primitive made the two
concerns independently observable and swappable.

\textbf{\texttt{explain} and \texttt{predict} carry no partition guard.}
Every other kernel primitive carries at least one partition guard:
\texttt{prepare} and \texttt{fit} require registered
\texttt{\{train,\ valid,\ dev\}}-tagged data; \texttt{evaluate} rejects
test-tagged and unregistered data; \texttt{assess} requires test-tagged
data and fires once per holdout. \texttt{explain} and \texttt{predict}
require only a fitted Model. Both may be called on any data, any number
of times, before or after \texttt{assess}. Explanation and prediction
are diagnostic, not part of the validity chain; the grammar has no
partition guard to place on them.

The \textbf{assess-once} constraint is a call-time guard, not a type
error: a second \texttt{assess} on the same holdout fails at runtime
(Section~\ref{sec-design-properties}). \texttt{fit} accepts
\texttt{\{train,\ valid,\ dev\}}-tagged data and handles preparation
internally per fold; calling \texttt{prepare} explicitly is available
but never required.

\textbf{The iterative/final \texttt{fit} distinction.} The type
signature
\texttt{fit\ ::\ (DataFrame\^{}g\ \textbar{}\ CVResult\ \textbar{}\ PreparedData)\ ×\ target\ →\ Model}
is identical whether \texttt{fit} is called during the iterate zone
(training a candidate model per fold, evaluated and discarded) or as the
final \texttt{fit(dev)} call after hyperparameters are fixed. These two
roles differ in commitment: per-fold models are expendable; the final
\texttt{fit(dev)} model is the one that proceeds to \texttt{assess}. The
grammar assigns both calls the same type because they satisfy the same
preconditions. The semantic distinction (which \texttt{fit} is final) is
invisible to the type system; the diagnostics layer catches it.

\textbf{Why 8?} The count is a design argument, not a proof: 8 is
sufficient for covering the supervised learning lifecycle without
collapsing a type boundary or losing a guard, and each primitive is
justified by what its removal would cost. Five primitives carry
load-bearing guards (\texttt{split}, \texttt{prepare}, \texttt{fit},
\texttt{evaluate}, \texttt{assess}); removing any of these drops a guard
the grammar relies on. The remaining three (\texttt{cv},
\texttt{predict}, \texttt{explain}) carry no guard but are structurally
necessary for typed composition: \texttt{cv} produces CVResult (without
it, fold rotation is hidden inside boundary logic and the type DAG
cannot express k-fold flow); \texttt{predict} is the only typed path
from Model to Predictions (without it, Predictions is unreachable as a
declared sort, which breaks stacking strategies that consume Predictions
and the diagnostics layer that operates on them); \texttt{explain}
produces Explanations (a terminal sort that cannot be substituted for
Metrics or Predictions). Whether 8 is \emph{minimal} in the formal sense
(no equivalent grammar exists with fewer primitives) is an open
question, a grammar that demoted \texttt{predict} to a strategy-level
operation rather than a primitive would lose typed composition for
Predictions but would not lose a guard. The claim here is that 8 is
sufficient and that each of the 8 has a structural function the others
cannot absorb without collapse, not that 8 is uniquely minimal.

\newpage

\subsection{The type DAG}\label{the-type-dag}

The primitives connect through a typed directed acyclic graph (Figure
1): DataFrame splits into train/valid/test partitions; \texttt{cv}
produces a rotation schedule (\(k\) train/validation index pairs)
consumed fold-by-fold by \texttt{fit}; iteration converges to terminal
\texttt{assess} on held-out test data. Stacking strategies (Appendix,
footnote ‡) introduce a controlled re-entry: out-of-fold predictions
reshape to a DataFrame for a second \texttt{fit}. The strategy
orchestration remains acyclic when unrolled into its own DAG; Figure 1
abstracts this path out.

\begin{figure}[H]
\vspace{-0.5em}
\centering\includegraphics[height=0.62\textheight,trim=0 30 0 50,clip]{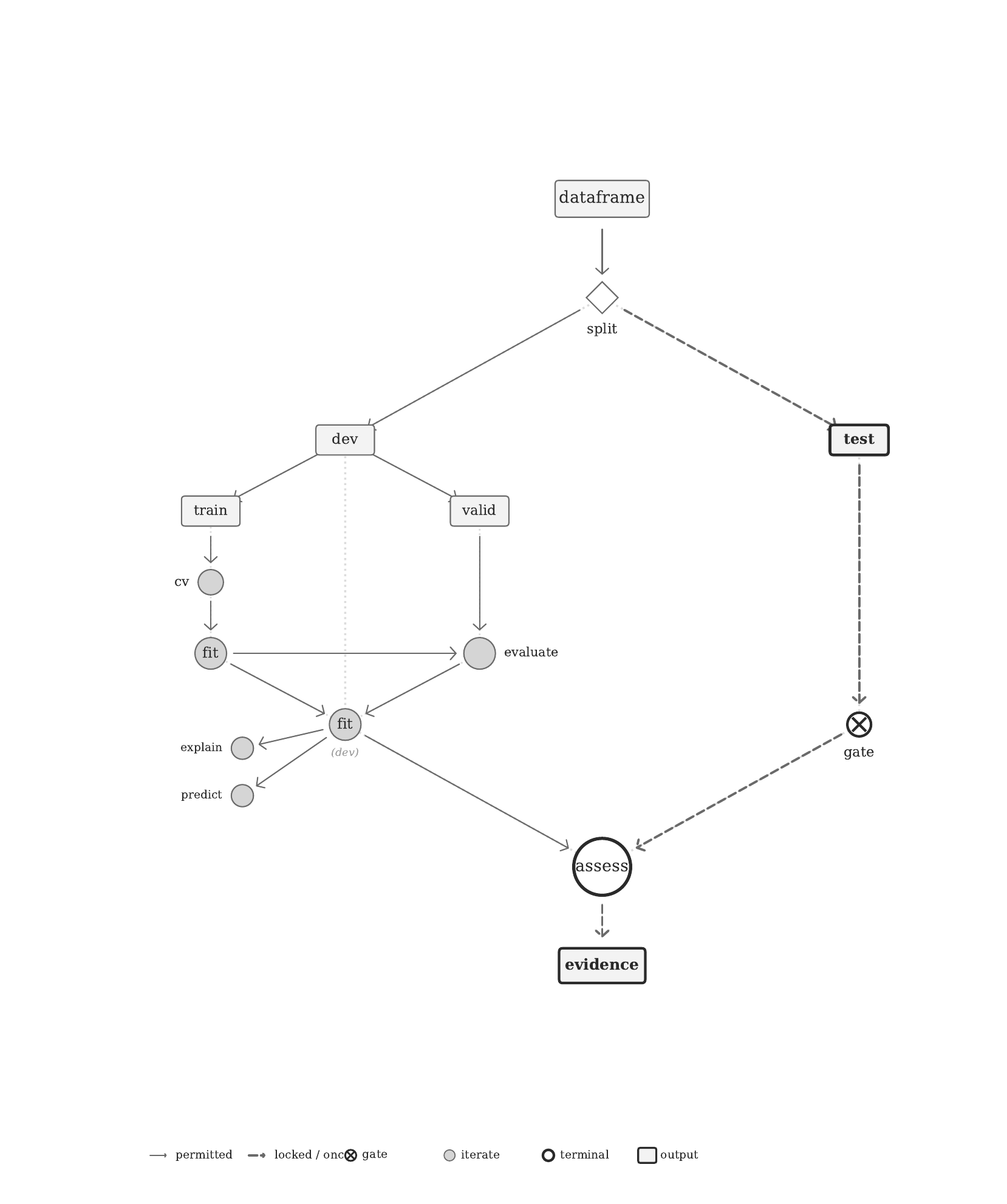}
\vspace{-0.5em}
\caption{The ML workflow grammar as a typed DAG for the base supervised workflow. Diamond: split decision producing dev and test partitions. \texttt{cv} transforms the training DataFrame into train/valid fold rotations. Solid arrows: permitted data flow. Dashed arrows: locked until terminal assessment. Gate ($\times$): test partition held until \texttt{assess} is called once, producing terminal evidence. The fit--evaluate loop iterates on train/valid folds; \texttt{fit(dev)} retrains on the full development set before the final commitment.}
\end{figure}

The Partition type exposes \texttt{.train}, \texttt{.valid},
\texttt{.test}, and \texttt{.dev} (train \(\cup\) valid).
Cross-validation (\citeproc{ref-stone1974cross}{Stone 1974};
\citeproc{ref-arlot2010survey}{Arlot and Celisse 2010}) is a separate
primitive (\texttt{cv}); Stone integrated cross-validatory choice with
predictive assessment, the framework on which the grammar's \texttt{cv}
operates. The primitive transforms a training DataFrame into a CVResult
with \texttt{.folds} and \texttt{.k}; the test partition stays on the
original Partition and is accessed at \texttt{assess} time. The full
workflow with cross-validation:

\begin{Shaded}
\begin{Highlighting}[]
\NormalTok{s }\OperatorTok{=}\NormalTok{ ml.split(df, target}\OperatorTok{=}\StringTok{"y"}\NormalTok{, seed}\OperatorTok{=}\DecValTok{42}\NormalTok{)}
\NormalTok{c }\OperatorTok{=}\NormalTok{ ml.cv(s.train, folds}\OperatorTok{=}\DecValTok{5}\NormalTok{, seed}\OperatorTok{=}\DecValTok{42}\NormalTok{)}
\NormalTok{model }\OperatorTok{=}\NormalTok{ ml.fit(c, }\StringTok{"y"}\NormalTok{, algorithm}\OperatorTok{=}\StringTok{"rf"}\NormalTok{, seed}\OperatorTok{=}\DecValTok{42}\NormalTok{)}
\NormalTok{metrics }\OperatorTok{=}\NormalTok{ ml.evaluate(model, s.valid)        }\CommentTok{\# iterate zone}
\NormalTok{best }\OperatorTok{=}\NormalTok{ ml.fit(s.dev, }\StringTok{"y"}\NormalTok{, algorithm}\OperatorTok{=}\StringTok{"rf"}\NormalTok{, seed}\OperatorTok{=}\DecValTok{42}\NormalTok{)}
\NormalTok{evidence }\OperatorTok{=}\NormalTok{ ml.assess(best, test}\OperatorTok{=}\NormalTok{s.test)      }\CommentTok{\# terminal, once}
\end{Highlighting}
\end{Shaded}

\texttt{cv} takes the training partition and creates a \(k\)-fold
rotation schedule; cross-validation is a training-set operation
(\citeproc{ref-hastie2009elements}{Hastie, Tibshirani, and Friedman
2009}, Ch. 7). The test rows remain on the originating partition,
untouched. \texttt{fit} consumes the CVResult and trains across folds;
\texttt{evaluate} measures on the held-out validation set (repeatable);
\texttt{assess} produces sealed Evidence on test data (terminal). A
second \texttt{assess} on the same holdout fails the guard. Preparation
is handled inside \texttt{fit} by default (per fold); calling
\texttt{prepare} explicitly is available for full control:

\begin{Shaded}
\begin{Highlighting}[]
\NormalTok{prepped }\OperatorTok{=}\NormalTok{ ml.prepare(s.train, target}\OperatorTok{=}\StringTok{"y"}\NormalTok{)  }\CommentTok{\# fit on train only}
\NormalTok{model }\OperatorTok{=}\NormalTok{ ml.fit(prepped, }\StringTok{"y"}\NormalTok{, algorithm}\OperatorTok{=}\StringTok{"rf"}\NormalTok{, seed}\OperatorTok{=}\DecValTok{42}\NormalTok{)}
\end{Highlighting}
\end{Shaded}

\textbf{Domain specializations.} The implementation exposes
\texttt{split}, \texttt{split\_temporal}, and \texttt{split\_group} as
independent entry points, each with domain-specific parameters and
guards. \texttt{split\_temporal} carries guards that \texttt{split} does
not: embargo enforcement, expanding-vs-sliding window policy, temporal
ordering with no future leakage. \texttt{split\_group} carries a
group-non-overlap guard that is meaningless for random splits. The same
pattern applies to \texttt{cv}, \texttt{cv\_temporal}, and
\texttt{cv\_group}. All specializations share the type signature
\texttt{DataFrame\ →\ Partition} (or \texttt{DataFrame\ →\ CVResult} for
cv variants); what differs is the guard system.

The specializations are constraint profiles, not type distinctions: each
conjugates the same grammatical verb for a different scientific domain.
Time series forecasting, where expanding-window validation prevents
future leakage (e.g., \citeproc{ref-ballarin2024reservoir}{Ballarin et
al. 2024}). For grouped data with repeated observations per unit,
subject-level splitting prevents within-unit leakage (e.g.,
\citeproc{ref-tampu2022inflation}{Tampu, Eklund, and Haj-Hosseini
2022}). Standard cross-sectional ML. The primitive count remains 8; the
API surface reflects that different fields require different structural
constraints on the same operation.

\subsection{The four hard constraints}\label{the-four-hard-constraints}

\begin{longtable}[]{@{}
  >{\raggedright\arraybackslash}p{(\linewidth - 8\tabcolsep) * \real{0.0500}}
  >{\raggedright\arraybackslash}p{(\linewidth - 8\tabcolsep) * \real{0.2200}}
  >{\raggedright\arraybackslash}p{(\linewidth - 8\tabcolsep) * \real{0.3800}}
  >{\centering\arraybackslash}p{(\linewidth - 8\tabcolsep) * \real{0.2200}}
  >{\centering\arraybackslash}p{(\linewidth - 8\tabcolsep) * \real{0.1300}}@{}}
\toprule\noalign{}
\begin{minipage}[b]{\linewidth}\raggedright
\#
\end{minipage} & \begin{minipage}[b]{\linewidth}\raggedright
Constraint
\end{minipage} & \begin{minipage}[b]{\linewidth}\raggedright
What it prevents
\end{minipage} & \begin{minipage}[b]{\linewidth}\centering
Leakage class
\end{minipage} & \begin{minipage}[b]{\linewidth}\centering
\(\Delta\)AUC
\end{minipage} \\
\midrule\noalign{}
\endhead
\bottomrule\noalign{}
\endlastfoot
1 & \textbf{Assess once per holdout} & Repeated test-set peeking & Class
II (\(d_z\) = 0.929) & +0.040 \\
2 & \textbf{Prepare after split, per fold} & Global preprocessing
leakage & Class I (\(d_z \approx 0\)) & \(\leq 0.005\) \\
3 & \textbf{Type-safe transitions} & Fitting on test/untagged data;
evaluating without a fitted model & Class II/III (\(d_z\) =
0.372--1.112) & +0.004--0.024 \\
4 & \textbf{No unregistered data into \texttt{fit}} & Feature selection
using test labels & Class II (\(d_z\) = 0.929) & +0.040 \\
\end{longtable}

Constraints 1 and 4 both target Class II leakage; they differ in
mechanism (per-holdout registry vs.~provenance gate at \texttt{fit}).
The companion study (\citeproc{ref-roth2026landscape}{Roth 2026})
measured the effects these constraints prevent: at the corpus median
(\(n \approx 1{,}900\)), peeking inflates by \(d_z\) = 0.929 and seed
cherry-picking by \(d_z\) = 0.885, enough to flip a model from ``below
threshold'' to ``deployable.'' Class III memorization effects (\(d_z\) =
0.372--1.112) scale with model capacity. Constraint 2 addresses an
effect that is empirically negligible (\(d_z \approx 0\)). But it costs
nothing (per-fold preparation is the default), and including it
preserves structural completeness across the four-class taxonomy.

Four rules. Codd needed twelve relational rules and eventually struggled
with the count. Chomsky's early programme required increasingly many
syntactic rules before principles-and-parameters reduced the apparatus.
The grammar chooses minimality: four constraints that cannot be
violated, everything else left to the implementation.

\subsection{Strategy families}\label{sec-strategy}

A practitioner does not call \texttt{split}, \texttt{fit},
\texttt{evaluate}, \texttt{assess} one at a time. They compose them into
workflows: screen several algorithms, tune the winner, stack an
ensemble, assess the result. The 8 primitives compose into 4 strategy
families that cover this full developmental cycle:

\begin{longtable}[]{@{}
  >{\raggedright\arraybackslash}p{(\linewidth - 4\tabcolsep) * \real{0.2000}}
  >{\raggedright\arraybackslash}p{(\linewidth - 4\tabcolsep) * \real{0.3500}}
  >{\raggedright\arraybackslash}p{(\linewidth - 4\tabcolsep) * \real{0.4500}}@{}}
\toprule\noalign{}
\begin{minipage}[b]{\linewidth}\raggedright
Family
\end{minipage} & \begin{minipage}[b]{\linewidth}\raggedright
Question
\end{minipage} & \begin{minipage}[b]{\linewidth}\raggedright
Strategies
\end{minipage} \\
\midrule\noalign{}
\endhead
\bottomrule\noalign{}
\endlastfoot
\textbf{Selection} & Which algorithm? & screen, compare, pick \\
\textbf{Optimization} & Which hyperparameters? & tune (grid, random,
bayesian) \\
\textbf{Evaluation} & How to rotate data? & kfold, grouped, temporal,
nested\_cv \\
\textbf{Ensemble} & How to combine models? & blend, bag, boost, stack \\
\end{longtable}

The strategies decompose into kernel primitives:

\begin{itemize}
\tightlist
\item
  \textbf{screen}: \(\forall\) algo: fit \(\to\) evaluate \(\to\) rank
  \(\to\) Leaderboard
\item
  \textbf{tune}: \(\forall\) params: fit \(\to\) evaluate \(\to\) pick
  \(\to\) TuningResult
\item
  \textbf{stack}: \(\forall\) model: fit \(\to\) out-of-fold predict
  \(\to\) fit\_meta \(\to\) StackedModel
\end{itemize}

No strategy requires a new primitive: each reduces to typed applications
of the 8 kernel verbs. That the core developmental workflow fits within
this decomposition (for the strategies examined here) is evidence that
the primitives capture structure, not an accidental API boundary. The
output containers Leaderboard, TuningResult, and StackedModel are
informal wrappers for composition results; they are not declared types
in the grammar's type system.

\textbf{The erased intermediate in \texttt{tune}.} The \texttt{tune}
decomposition hides a type the grammar does not declare: after
\texttt{pick} identifies optimal parameters, those parameters (a
\texttt{HyperParameters} intermediate) feed the final \texttt{fit(dev)}
call. The grammar's visible type chain is
\texttt{TuningResult\ →\ fit\ →\ Model}; the \texttt{HyperParameters}
intermediate exists in every implementation but is not a sort in the
type DAG. The compression is intentional: \texttt{tune} is a strategy,
not a primitive, and its internal types are outside the grammar's
minimal vocabulary. A conforming extension could expose
\texttt{HyperParameters} as a declared sort with its own typed accessor
from \texttt{TuningResult}, making the refitting step explicit.

\textbf{Grammar predictions.} The constraints and composition rules
generate testable predictions: if a strategy violates a constraint,
measurable inflation should follow. Three were tested in the companion
study before observing results:

\begin{longtable}[]{@{}
  >{\raggedright\arraybackslash}p{(\linewidth - 4\tabcolsep) * \real{0.4000}}
  >{\raggedright\arraybackslash}p{(\linewidth - 4\tabcolsep) * \real{0.3000}}
  >{\raggedright\arraybackslash}p{(\linewidth - 4\tabcolsep) * \real{0.3000}}@{}}
\toprule\noalign{}
\begin{minipage}[b]{\linewidth}\raggedright
Prediction
\end{minipage} & \begin{minipage}[b]{\linewidth}\raggedright
Source
\end{minipage} & \begin{minipage}[b]{\linewidth}\raggedright
Result
\end{minipage} \\
\midrule\noalign{}
\endhead
\bottomrule\noalign{}
\endlastfoot
Screen inflation (\(d_z > 0\)): selecting best-of-\(K\) algorithms
inflates performance & Selection family; consistent with Cawley and
Talbot (\citeproc{ref-cawley2010overfitting}{2010}) & Confirmed: \(d_z\)
= +0.27, \(K\)-invariant \\
Stack leakage (\(d_z > 0\)): out-of-fold meta-learner leaks through fold
labels & Ensemble family & \textbf{Falsified}: \(d_z\) = -0.22;
\texttt{stack()} is empirically safe \\
Seed inflation (\(d_z > 0\)): reporting best-of-\(S\) seeds inflates
performance & Assess-once constraint & Confirmed: \(d_z\) = +0.89,
prevalence 92\% \\
\end{longtable}

Two of three confirmed; one falsified. The confirmed predictions were
not equally risky: screen inflation was consistent with Cawley and
Talbot (\citeproc{ref-cawley2010overfitting}{2010})'s prior findings on
selection bias, so its confirmation was expected. Seed inflation was the
more novel claim. The falsified prediction (stack leakage) was
structurally safe to lose: the grammar's correctness does not depend on
stacking being leaky. The falsification is evidence that the grammar
generates specific, wrong-able hypotheses, not post-hoc rationalization.
The full experimental results are in Section~\ref{sec-empirical}.

The grammar generates further predictions. The companion study already
provides partial evidence for some (nested CV bias is measured
indirectly through the HP tuning experiment; the high-capacity ordering
NB \textless{} LR \textless{} XGB \textless{} RF \textless{} KNN
\textless{} DT is confirmed at duplication, with neural-network
extension predicted but unmeasured in both papers). Two predictions
remain genuinely open: (1) a between-subjects experiment should show
that the grammar's structural enforcement reduces leakage rates in
human-written code compared to sklearn: if it does not, the grammar is
correct but practically inert; (2) the grammar should produce the same
leakage reduction for LLM-generated code as for human-written code: if
the structural guarantee holds, the generation source (human or LLM) is
irrelevant.

\subsection{The terminal boundary}\label{the-terminal-boundary}

In a foundational essay on evaluation methodology, Scriven
(\citeproc{ref-scriven1967methodology}{1967}) coined \emph{formative}
and \emph{summative} evaluation, drawing a line that education research
has respected ever since: \emph{formative} evaluation improves the thing
being developed; \emph{summative} evaluation judges the finished
product. The two serve different purposes and must not contaminate each
other. ML has the same distinction and has lost it. In sklearn,
\texttt{cross\_val\_score} reports a quantity called
\texttt{test\_score} that is formative feedback, not a terminal test. In
tidymodels, the ``assessment set'' is a held-out CV fold used for
iteration, not for judgment. When the same word covers both, nothing in
the vocabulary marks the boundary, and 648 papers crossed it.

The grammar recovers the distinction as a type boundary.
\texttt{evaluate} is the homework: formative, repeatable, safe to learn
from. \texttt{assess} is the exam: summative, once, irreversible. The
type system is the wall between them, enforced at call time.

\begin{itemize}
\tightlist
\item
  \textbf{Evaluate} operates on validation data. It is repeatable,
  cheap, and safe for decisions. It informs the iterate cycle: the
  practitioner reads the Metrics output, adjusts, and fits again.
  Metrics is terminal in the type system (no primitive accepts it as
  input), but it drives iteration at the practitioner level.
\item
  \textbf{Assess} operates on test data. It is terminal, once per
  holdout, and irreversible. It is the commit: the model's performance
  on data it has never been influenced by or seen.
\end{itemize}

The terminal boundary prevents accumulation of selection pressure. It
does not eliminate per-decision noise: at practical dataset sizes,
peeking-style selection over correlated draws carries an expected
inflation that decays as \(1/\sqrt{n}\) but retains a diversity residual
at large \(n\); seed-style selection over independent draws is purely
noise exploitation and vanishes by \(n \geq 5{,}000\)
(\citeproc{ref-roth2026landscape}{Roth 2026}). This inflation grows
logarithmically in \(K\) (the companion study fits \(\log K\) at
\(R^2 > 0.99\); see Roth (\citeproc{ref-roth2026landscape}{2026}),
Limitation 11). Varma and Simon (\citeproc{ref-varma2006bias}{2006})
provide independent corroboration on null data (\(n = 40\) samples,
class labels randomly assigned so true error is 50\%): single-loop CV
used for both hyperparameter tuning and error estimation produces
8--12pp optimistic bias (reported error 37.8--41.7\% against the 50\%
null), smaller than a naive best-of-\(K\) extrapolation because grid
points in their \((C, \gamma)\) search are correlated, reducing
effective \(K\); nested CV mitigates the bias to \(\leq 4\)pp.~The
grammar's contribution is making the summative decision the \emph{only}
decision allowed to touch the test set within a given research cycle,
bounding the total selection pressure to one terminal query rather than
letting it accumulate across an uncontrolled sequence of evaluations.

The sample-splitting logic is textbook
(\citeproc{ref-hastie2009elements}{Hastie, Tibshirani, and Friedman
2009, chap. 7}): hold out a test set for generalization estimation
(§7.2), use cross-validation for model selection (§7.10), and never
preprocess before splitting (§7.10.2, ``the wrong way''). The grammar
adds no new statistical insight: Hastie's textbook already spells the
three-tier hierarchy out, and Bates, Hastie, and Tibshirani
(\citeproc{ref-bates2024crossvalidation}{2024}) sharpened what cv error
actually estimates. The grammar's contribution is translating the
textbook advice into a rejection criterion that fires at call time.
Three feedback layers, three typed boundaries: cv-aggregated error
inside \texttt{fit} drives hyperparameter tuning across folds of
\texttt{s.train} (the Bates \(\mathrm{Err}\) quantity, a procedure-level
estimate); \texttt{evaluate(model,\ s.valid)} produces a
single-realization \(\mathrm{Err}_{XY}\) sample estimate on the separate
valid partition for model selection and steering between iterations;
\texttt{assess(model,\ s.test)} produces the terminal
\(\mathrm{Err}_{XY}\) sample estimate, the once-only commitment. The two
textbook failure modes (preprocessing before splitting, §7.10.2; reusing
the test set, the ``vault'' passage at §7.2) map to Constraints 4, 2,
and 1. The practitioner-side confusion (cv folds doubling as
model-selection signal, s.valid omitted, ``cv score'' read as a
deployment guarantee) comes from collapsing the three tiers in the
default sklearn workflow; the grammar refuses the collapse by typing the
boundaries separately. Treating any of these finite-sample estimates as
a population-level guarantee is a category error that the type
distinction makes visible: Metrics (formative, repeatable) versus
Evidence (summative, terminal).

Bates et al.'s analysis operates entirely within the evaluate zone and
makes no claim about terminal holdout assessment. The Metrics/Evidence
type distinction is this paper's extension of their finding, not their
conclusion. The grammar provides the complementary formalization for the
terminal boundary: the evaluate/assess distinction, implicit in every
textbook's sample-splitting advice, is enforced by the types, not by the
student's memory.

\section{Design Properties}\label{sec-design-properties}

The grammar has three design properties. Each can be checked by anyone
reimplementing the specification. They are design constraints, not
theorems, holding by construction rather than by proof.

\textbf{Formal characterization.} The system uses call-time guards in
the spirit of typestate (\citeproc{ref-strom1986typestate}{Strom and
Yemini 1986}): what you can do next depends on where you are. The
assess-once constraint is genuine typestate (a specific holdout
transitions from unassessed to assessed); the broader workflow (where
\texttt{split} produces a Partition and \texttt{fit} produces a distinct
Model) is closer to a session protocol than per-object typestate. The
term ``grammar'' is used in the Wilkinson--Wickham sense (a
decomposition of a domain into primitives with composition rules and a
rejection criterion), not Chomsky's generative sense.

The type system has four states. Each primitive either advances the
state, rejects the call, or does not apply. The transition table makes
the rejection criterion visible at a glance: every ``reject'' cell is a
leakage vector that the grammar closes.

\begin{table}[H]
\centering
\footnotesize
\setlength{\tabcolsep}{3pt}
\begin{tabular}{l c c c c c c c c}
\toprule
\textbf{State} & \texttt{split} & \texttt{cv} & \texttt{prep.} & \texttt{fit} & \texttt{eval.} & \texttt{expl.} & \texttt{pred.} & \texttt{assess} \\
\midrule
Untagged    & Partit. & rej. & rej.       & rej.   & rej.    & rej.        & rej.      & rej. \\
Partitioned & -       & CVRes. & PrepData & Fitted & rej.    & rej.        & rej.      & rej. \\
Fitted      & -       & -    & -          & -      & Metrics & Explan.     & Predict.  & Evidence \\
Assessed    & -       & -    & -          & -      & Metrics & Explan.     & Predict.  & rej. \\
\bottomrule
\end{tabular}
\end{table}

States correspond to the data/model lifecycle: Untagged (raw DataFrame),
Partitioned (after \texttt{split}), Fitted (after \texttt{fit}),
Assessed (after \texttt{assess}). ``reject'' means the call fails with
an error; a dash in the table means the verb does not apply to that
state's type. \texttt{explain} and \texttt{predict} remain available
after assessment; \texttt{assess} fires once per holdout (Fitted →
Assessed) and rejects any subsequent call on the same test partition.

The grammar says no in two ways. Structural rejection is static: a
function that accepts Model cannot receive a Partition; the types are
incompatible and the call never begins. Guard rejection is dynamic: the
types match but the provenance registry remembers that this test
partition has already been assessed, and the guard fires before the call
completes. The four hard constraints are four policies enforced by one
mechanism (type checking + guard checking), each targeting a distinct
leakage vector (three address Class II through different mechanisms; one
addresses Class I). Assess-once is a guard; the type DAG prevents
nonsensical type chains structurally. The policies are independent (each
addresses a specific leakage vector); the enforcement mechanism is
shared. Full algebraic formalization of the guard system is future work.

\subsection{Connectivity (not algebraic
closure)}\label{connectivity-not-algebraic-closure}

A grammar with dead-end types forces the practitioner to leave the type
system and work around it. This grammar has no dead ends: every type in
the DAG is reachable from DataFrame through the primitives. The
non-terminal types form three reachable paths from DataFrame: a direct
path (DataFrame \(\to\) Partition \(\to\) Model, via \texttt{fit}), an
explicit-prepare path (DataFrame \(\to\) Partition \(\to\) PreparedData
\(\to\) Model, via \texttt{prepare} then \texttt{fit}), and a CV branch
(Partition \(\to\) DataFrame\(_{\text{train}}\) \(\to\) CVResult \(\to\)
Model, via fold-aware \texttt{fit}). The direct path is the default;
explicit preparation is optional. The terminal types (Predictions,
Metrics, Evidence, Explanation) are all reachable from Model. No type is
orphaned; any conforming extension must preserve this connectivity.

The grammar is not closed in Codd's sense. In the relational algebra,
every operator returns a relation, so operators compose freely. Here,
\texttt{predict} returns Predictions, not a DataFrame; you cannot call
\texttt{split} on a prediction. The types that prevent leakage are the
same types that prevent free composition. A fully composable grammar
would let you pipe anything into anything, including leaky workflows.
Free composition is not the design goal; correctness is.

\subsection{Generativity}\label{generativity}

A grammar that limits how many models you can try before committing
would be intolerable. This grammar does not: split → (fit →
evaluate)\(^n\) → assess is valid for any \(n\). Each \texttt{fit} and
\texttt{evaluate} call operates on the original Partition. The Metrics
value returned by \texttt{evaluate} is terminal in the type DAG (no
primitive accepts Metrics as a data input), but it is read by the
practitioner (and by the \texttt{tune}/\texttt{pick} strategies that
wrap the primitives) to choose the next call's hyperparameters. The data
flow stays anchored in Partition; the information flow through Metrics
is operator-mediated reconfiguration, not Metrics-as-data threaded into
the next \texttt{fit}. The cycle exists in the workflow pattern, not in
the linear type chain. A weak form of generativity; any system with a
loop generates an infinite language. The claim is not about
expressiveness class (the grammar's formal language-theoretic
classification is an open question) but about practical scope: the
grammar does not artificially restrict how many models a practitioner
trains before committing to assessment.

\textbf{Scope.} The grammar covers batch supervised learning on a
complete dataset available at split time, including automatic production
retrain cycles. Multi-task learning (multiple simultaneous targets),
self-supervised learning (no explicit labels), few-shot learning (\(N\)
too small to partition meaningfully), federated learning (data never
centralized), and online learning (incremental data arrival) each
violate at least one foundational assumption and fall outside the
grammar's scope. The grammar permits all valid workflows within its
scope (no valid workflow is rejected) and makes no claim about valid ML
workflows in general.

\subsection{Rejection criterion}\label{rejection-criterion}

Here is where the grammar earns its keep. Each row below is a workflow
that looks reasonable, runs without error in sklearn, and produces a
number that is wrong. The grammar rejects all of them:

\begin{longtable}[]{@{}
  >{\raggedright\arraybackslash}p{(\linewidth - 4\tabcolsep) * \real{0.4000}}
  >{\raggedright\arraybackslash}p{(\linewidth - 4\tabcolsep) * \real{0.4000}}
  >{\centering\arraybackslash}p{(\linewidth - 4\tabcolsep) * \real{0.2000}}@{}}
\toprule\noalign{}
\begin{minipage}[b]{\linewidth}\raggedright
Invalid workflow
\end{minipage} & \begin{minipage}[b]{\linewidth}\raggedright
Why rejected
\end{minipage} & \begin{minipage}[b]{\linewidth}\centering
Leakage class
\end{minipage} \\
\midrule\noalign{}
\endhead
\bottomrule\noalign{}
\endlastfoot
assess \(\to\) assess (same holdout) & Terminal: once per holdout &
Class II \\
prepare(all) \(\to\) split \(\to\) fit & Prepare must follow split &
Class I \\
select\_features(X, y) \(\to\) split \(\to\) fit & No unregistered data
into \texttt{fit} & Class II \\
evaluate(data) without prior fit & Type error: no Model & Class III \\
prepare(test) & Guard: test tag not in \{train, valid, dev\} & Class
I \\
fit(test) & Guard: test tag not in \{train, valid, dev\} & Class III \\
evaluate(test) & Guard: test tag reserved for assess & Class II \\
\end{longtable}

Concretely, an invalid workflow raises at the verb boundary:

\begin{Shaded}
\begin{Highlighting}[]
\NormalTok{s }\OperatorTok{=}\NormalTok{ ml.split(df, target}\OperatorTok{=}\StringTok{"y"}\NormalTok{, seed}\OperatorTok{=}\DecValTok{42}\NormalTok{)}
\NormalTok{m }\OperatorTok{=}\NormalTok{ ml.fit(s.dev, }\StringTok{"y"}\NormalTok{, algorithm}\OperatorTok{=}\StringTok{"rf"}\NormalTok{, seed}\OperatorTok{=}\DecValTok{42}\NormalTok{)}
\NormalTok{e1 }\OperatorTok{=}\NormalTok{ ml.assess(m, test}\OperatorTok{=}\NormalTok{s.test)   }\CommentTok{\# succeeds}
\NormalTok{e2 }\OperatorTok{=}\NormalTok{ ml.assess(m, test}\OperatorTok{=}\NormalTok{s.test)   }\CommentTok{\# fails}
\CommentTok{\# PartitionError: test holdout already assessed}
\end{Highlighting}
\end{Shaded}

\textbf{Assess-once (by construction).} Constraint 1 specifies: for test
partition \(t\), assess(\(m\), \(t\)) is valid if and only if \(t\) has
not been previously assessed in the provenance registry, regardless of
which model. After the first call, \(t\) is marked as assessed; any
subsequent call on the same holdout (same model or different) fails the
guard. The implementations additionally set \(m\).assessed \(\gets\)
true as a fast-path diagnostic, but the per-holdout check is the
structurally load-bearing enforcement. Similarly, prepare(all\_data)
before split violates Constraint 2 (the data is unregistered in the
provenance registry, so the guard rejects it), and select\_features(X,
y) before split violates Constraint 4. All three are rejected at call
time. These are not theorems; they are consequences of the type guards.
The claim is that any conforming implementation must enforce them. The
Appendix (§ Invalid workflows as guard failures) walks through each
entry in the table above.

This grammar is a pure rejection function, not a generation function:
its value lies not in the workflows it enables but in the workflows it
rejects at call time.

The grammar's rejection happens at \textbf{call time}, not at
construction time. Chomsky's grammar makes certain strings underivable;
Codd's constraints make certain rows unstorable; the ML grammar makes
certain calls non-executable: three flavors of structural rejection
distinguished by enforcement layer. \texttt{ml.fit(s.test,\ "y")} is a
syntactically valid Python expression, but it raises
\texttt{PartitionError} before any computation begins: the invalid
workflow is writable in the host language but not executable through the
grammar's API. This is the price of embedding a grammar in a dynamically
typed host: the grammar cannot prevent you from writing the line, only
from running it. Runtime guards at every verb entry point are the
strongest enforcement available without a native partition type in the
host language.

The grammar's \texttt{partition\_tag} functions as an information flow
label (\citeproc{ref-myers1999jflow}{Myers 1999}): train, valid, and
test are security levels, and the guards enforce that test-labeled data
cannot flow into training operations. The grammar's version is simpler
than JFlow's: it checks labels at primitive boundaries, not through
every line of code. Same idea applied at a different boundary:
train/test instead of public/private. Drobnjaković, Subotić, and Urban
(\citeproc{ref-drobnjakovic2024abstract}{2024})'s abstract
interpretation approach arrives at the same insight from the static
analysis side; the grammar enforces it dynamically.

All four rules are purely negative: they define the boundary of valid
workflows without ranking or recommending within it. The grammar does
not specify which algorithm to choose, how many folds to use, or when to
stop iterating; those are semantics-layer decisions. The positive
structure of a good ML workflow is the diagnostics and strategy layer's
job, not the grammar's.

To my knowledge, no existing supervised-learning framework documented in
the peer-reviewed literature enforces a terminal assess-once constraint
at call time, rejecting the violation rather than detecting it
afterward. The grammar does not lint completed workflows; it rejects
invalid ones as they are constructed. This claim is bounded to the
grammar's own type system: operations outside the grammar's eight
primitives can always produce leakage that the grammar cannot see.
Adjacent tooling, experiment-tracking and ML-Ops systems (MLflow,
Kubeflow Pipelines, DVC, ZenML), provides typed pipeline steps, model
registries with stage transitions, and audit trails, but none of these
enforce a \emph{call-time} terminal-assessment boundary on test-set
access at the language-API level; their guarantees are post-hoc auditing
and workflow orchestration rather than runtime rejection of an invalid
evaluate-after-assess sequence. A complete survey of every ML/MLOps
system is outside scope. The closest call-time holdout-protection
mechanism in the literature is Thresholdout
(\citeproc{ref-dwork2015reusable}{Dwork et al. 2015}), which enforces
adaptive-query bounds via differential privacy and permits
\emph{multi-query} reuse up to a DP-bounded information leak budget.
Thresholdout is a relaxation of assess-once (permitting multi-query
reuse under a privacy budget), not a strengthening. Both mechanisms
operate at call time; they target different scales: terminal assess-once
for within-session zero-reuse, Thresholdout for adaptive multi-query
reuse across sessions.

The workflow's structural guarantees are a minimum over its constituent
primitives: a violation at \texttt{split()} propagates through all
downstream operations, and no downstream primitive recovers correctness
from an upstream violation. Hence the four hard constraints are placed
at \texttt{split} and \texttt{prepare} rather than at \texttt{assess}:
the terminal boundary is too late to undo upstream contamination.

\section{Embedding in Code}\label{embedding-in-code}

\subsection{Implementation}\label{implementation}

The grammar is implemented in two languages: Python (\texttt{mlw} on
PyPI, imported as \texttt{import\ ml}) and R (\texttt{ml} on CRAN). Both
expose all 8 kernel primitives with identical type signatures:

\begin{Shaded}
\begin{Highlighting}[]
\CommentTok{\# Python}
\ImportTok{import}\NormalTok{ ml}

\NormalTok{s }\OperatorTok{=}\NormalTok{ ml.split(df, target }\OperatorTok{=} \StringTok{"y"}\NormalTok{, seed }\OperatorTok{=} \DecValTok{42}\NormalTok{)}
\NormalTok{model }\OperatorTok{=}\NormalTok{ ml.fit(s.train, }\StringTok{"y"}\NormalTok{, algorithm }\OperatorTok{=} \StringTok{"rf"}\NormalTok{, seed }\OperatorTok{=} \DecValTok{42}\NormalTok{)}
\NormalTok{metrics }\OperatorTok{=}\NormalTok{ ml.evaluate(model, s.valid)}
\NormalTok{final }\OperatorTok{=}\NormalTok{ ml.assess(model, test}\OperatorTok{=}\NormalTok{s.test)}
\end{Highlighting}
\end{Shaded}

\begin{Shaded}
\begin{Highlighting}[]
\CommentTok{\# R}
\FunctionTok{library}\NormalTok{(ml)}

\NormalTok{s }\OtherTok{\textless{}{-}} \FunctionTok{ml\_split}\NormalTok{(df, }\AttributeTok{target =} \StringTok{"y"}\NormalTok{, }\AttributeTok{seed =} \DecValTok{42}\NormalTok{)}
\NormalTok{model }\OtherTok{\textless{}{-}} \FunctionTok{ml\_fit}\NormalTok{(s}\SpecialCharTok{$}\NormalTok{train, }\StringTok{"y"}\NormalTok{, }\AttributeTok{algorithm =} \StringTok{"rf"}\NormalTok{, }\AttributeTok{seed =} \DecValTok{42}\NormalTok{)}
\NormalTok{metrics }\OtherTok{\textless{}{-}} \FunctionTok{ml\_evaluate}\NormalTok{(model, s}\SpecialCharTok{$}\NormalTok{valid)}
\NormalTok{final }\OtherTok{\textless{}{-}} \FunctionTok{ml\_assess}\NormalTok{(model, }\AttributeTok{test =}\NormalTok{ s}\SpecialCharTok{$}\NormalTok{test)}
\end{Highlighting}
\end{Shaded}

The portability claim rests on two reference implementations (Python and
R) satisfying the same conformance conditions across language
boundaries. The specification in the Appendix is the authoritative
reference; conformance is defined behaviorally, not numerically:
\emph{any implementation that satisfies the conformance conditions below
is conforming regardless of host language}. Cross-language numerical
agreement is not part of the spec: a Python
\texttt{fit(...,\ algorithm="rf",\ seed=42)} and an R
\texttt{ml\_fit(...,\ algorithm="rf",\ seed=42)} call different
underlying implementations (e.g., scikit-learn vs.~ranger) whose random
number generation, splitting criteria, and defaults differ; identical
AUCs across languages are not expected and not required. The convergence
the two reference implementations demonstrate is structural (same types,
same guards, same conformance verdicts), not numerical.

\subsection{Conformance conditions}\label{conformance-conditions}

Codd (\citeproc{ref-codd1970relational}{1970}) defined the relational
model as a mathematical specification; the later ``Codd rules''
formalized a practical conformance standard against it. I propose
conformance conditions in the same spirit, but the analogy is limited:
Codd's anchor was mathematical (relational completeness); the grammar's
is empirical (leakage effect sizes that justify which constraints
exist). The eight conditions below test whether an implementation does
what the specification says it should, not whether the specification is
complete.

An implementation is conforming if and only if it satisfies all eight.
Conditions (1)--(4), (6), and (8) are testable from the API surface
alone; conditions (5) and (7) require additional support discussed
below:

\begin{enumerate}
\def\labelenumi{\arabic{enumi}.}
\tightlist
\item
  \texttt{split} produces a Partition with tagged train/valid/test/dev
  DataFrames
\item
  \texttt{fit} requires a registered DataFrame with
  \texttt{partition\_tag} \(\in\) \texttt{\{train,\ valid,\ dev\}};
  unregistered or test-tagged data raises \texttt{PartitionError}
\item
  \texttt{evaluate} and \texttt{assess} require a fitted Model
\item
  \texttt{assess} requires test-tagged data that has not been previously
  assessed; after the call, the holdout is marked as assessed and any
  subsequent call on the same partition fails (regardless of model
  identity: the registry tracks partition fingerprint, not model)
\item
  In declarative mode, \texttt{fit} applies \texttt{prepare}
  independently per fold; the scaler/encoder state from fold \(k\) does
  not leak into fold \(k+1\)
\item
  Evidence is a named class not substitutable for Metrics in the host
  language; \texttt{isinstance(result,\ Evidence)} is distinguishable
  from \texttt{isinstance(result,\ Metrics)}
\item
  \texttt{cv} produces a CVResult that blocks direct access to
  \texttt{.train}, \texttt{.valid}, \texttt{.test}, and \texttt{.dev};
  the test partition remains on the originating Partition
\item
  \texttt{evaluate} rejects test-tagged data; test data is reserved for
  \texttt{assess}
\end{enumerate}

The practical value of the conformance conditions is decidability:
without a grammar, ``does this workflow have structural leakage?'' has
no principled answer. With it, structural correctness is checkable in
constant time: eight primitives in the type DAG, four hard constraints,
eight conformance conditions. Methodological adequacy (algorithm choice,
metric selection, sample size) remains a judgment call. The test
produces a binary verdict, not an opinion.

Six of the eight conditions are verifiable from the API surface alone:
\texttt{fit} rejects unregistered data with a \texttt{PartitionError}
whose message directs the user to \texttt{split} first; \texttt{cv}
returns a CVResult that raises on partition access;
\texttt{evaluate}/\texttt{assess} reject mistyped inputs at call time.
The closed-world limitation of prior versions (where untagged data
passed silently) is closed.

Two conditions require additional support. Condition (5) (per-fold
isolation of \texttt{prepare} state) cannot be verified by external API
calls alone; the verification path is either reference-implementation
inspection (the Python and R implementations expose deterministic state
digests per fold for this purpose) or an adversarial probe, a synthetic
dataset on which non-isolated state would produce a different fitted
model than isolated state. Condition (7) (CVResult attribute blocking)
is testable in the obvious sense (attribute access raises), but the
protection it offers is bounded: it prevents accidental access
\emph{through the CVResult}, not access to the originating Partition
that the user already holds. The conformance condition rejects the
failure mode it is designed to catch (a user accidentally pulling
\texttt{cv\_result.test}); workflows that bypass CVResult entirely
(e.g., manually holding \texttt{s.test} and threading it into the wrong
primitive) are caught by conditions (4) and (8) instead. Where the
umbrella conformance claim depends on API-surface testability, it does
so over conditions (1)--(4), (6), (8), with (5) and (7) carrying the
caveats noted here.

\section{Comparison with Existing
Frameworks}\label{comparison-with-existing-frameworks}

The grammar claims to do something that, to my knowledge, no existing
framework does. The table below tests this claim against the three
closest alternatives on the four dimensions that matter for leakage
prevention:

\begin{longtable}[]{@{}
  >{\raggedright\arraybackslash}p{(\linewidth - 8\tabcolsep) * \real{0.3548}}
  >{\centering\arraybackslash}p{(\linewidth - 8\tabcolsep) * \real{0.1613}}
  >{\centering\arraybackslash}p{(\linewidth - 8\tabcolsep) * \real{0.1613}}
  >{\centering\arraybackslash}p{(\linewidth - 8\tabcolsep) * \real{0.1613}}
  >{\centering\arraybackslash}p{(\linewidth - 8\tabcolsep) * \real{0.1613}}@{}}
\toprule\noalign{}
\begin{minipage}[b]{\linewidth}\raggedright
Framework
\end{minipage} & \begin{minipage}[b]{\linewidth}\centering
Per-fold prep
\end{minipage} & \begin{minipage}[b]{\linewidth}\centering
Terminal assess
\end{minipage} & \begin{minipage}[b]{\linewidth}\centering
Provenance
\end{minipage} & \begin{minipage}[b]{\linewidth}\centering
Reject at call time
\end{minipage} \\
\midrule\noalign{}
\endhead
\bottomrule\noalign{}
\endlastfoot
\textbf{sklearn} & via Pipeline\(^p\) & --- & --- & --- \\
& & & & \\
\textbf{tidymodels} & via recipes\(^p\) & convention\(^l\) & --- &
--- \\
& & & & \\
\textbf{mlr3} & via PipeOps\(^p\) & --- & --- & --- \\
& & & & \\
\textbf{ml (this grammar)} & default & enforced\(^b\) &
content-addressed & \textbf{yes} \\
\end{longtable}

\(^p\)Inside the framework's pipeline/workflow/PipeOps abstraction only;
manual preprocessing outside leaks silently. \(^l\)\texttt{last\_fit()}
is a convention that encourages one-shot use but does not guard against
bypass: a practitioner can re-fit and re-evaluate on the test set
without runtime error. \(^b\)\texttt{assess()} rejects re-call on the
same holdout at runtime; on by default; \texttt{config(guards="off")}
disables all guards. Operations outside the grammar's eight primitives
(e.g., raw sklearn preprocessing) are not covered regardless. The
difference is default posture: the grammar defaults to enforcement and
requires an explicit exit; sklearn, tidymodels, and mlr3 default to
flexibility.

The grammar leads on methodological enforcement (the four columns above)
but trails on ecosystem breadth, maturity, and arbitrary code coverage.
sklearn's 17-year ecosystem and tidymodels' integration with the R
tidyverse are strengths the grammar does not replicate.

\subsection{tidymodels and caret}\label{tidymodels-and-caret}

Kuhn and Silge (\citeproc{ref-kuhn2022tidy}{2022}) is the closest
existing framework. Its recipes enforce per-fold preprocessing when used
inside workflows, preventing Class I leakage structurally. It deserves
explicit credit as the most important prior work. tidymodels itself
succeeded the earlier \texttt{caret} package
(\citeproc{ref-kuhn2008caret}{Kuhn 2008}), which introduced unified
resampling and tuning interfaces in R but offered no structural guard
against re-using the test set; tidymodels added recipe-based per-fold
preprocessing and \texttt{last\_fit()} on top of that lineage.

The grammar inherits tidymodels' insight that preprocessing must be
per-fold, and extends it in one direction: terminal assessment. This is
a design tradeoff, not a deficit in tidymodels: tidymodels optimizes for
flexibility and ecosystem integration within R; the grammar optimizes
for structural enforcement at the cost of ecosystem breadth. tidymodels
types are implementation-level (recipe, parsnip model, workflow) and
ensure that preprocessing steps are applied consistently.
\texttt{last\_fit()} is the closest existing analogue to the grammar's
assess-once guard: it encourages a one-shot terminal evaluation on the
test set after model selection is complete. But \texttt{last\_fit()} is
a recommended workflow function, not a guard; the user can bypass it and
call \texttt{collect\_metrics()} on test data repeatedly without raising
an error. No type marks a model as assessed. The grammar closes that gap
with a terminal type distinction that makes repeated test-set assessment
a runtime failure. The two frameworks agree on what the correct workflow
is; they differ on whether the incorrect one should be discouraged or
rejected.

\subsection{D3M and AutoML grammars}\label{d3m-and-automl-grammars}

DARPA's D3M programme developed a library of typed ML primitives with
automated pipeline search (\citeproc{ref-smith2020mlbazaar}{Smith et al.
2020}); Drori et al. (\citeproc{ref-drori2021alphad3m}{2021}) extends
this with meta-reinforcement-learning pipeline synthesis (LSTM sequence
model + Monte Carlo tree search, AlphaZero-style). Industrial AutoML
systems including auto-sklearn
(\citeproc{ref-feurer2015autosklearn}{Feurer et al. 2015}) take a
different route: a Bayesian-optimization search over a fixed
scikit-learn-shaped hypothesis space, with meta-learning warm-starts and
ensemble construction. In every case, the typed pipeline structure
exists to \emph{constrain the search}: a D3M- or auto-sklearn-generated
pipeline can still reuse the test set or cherry-pick seeds because the
type system answers ``what pipelines can I build?''; mine answers ``what
workflows are valid?''

\subsection{NBLyzer and abstract
interpretation}\label{nblyzer-and-abstract-interpretation}

Drobnjaković, Subotić, and Urban
(\citeproc{ref-drobnjakovic2024abstract}{2024}) applied abstract
interpretation, a principled framework from programming language
semantics, to ML leakage detection in Jupyter notebooks. Their system
propagates partition-membership labels (Train, Test, Full, Unknown)
through data-flow operations and flags any Test-labeled value flowing
into a fit-class operation as a leakage error. Evaluated on 2,111 Kaggle
notebooks, the approach achieved 93\% precision. This is the closest
existing work to a formal treatment of pipeline correctness.

The tradeoff is scope versus timing. Their tool examines completed code
and catches violations the grammar cannot see: arbitrary Python, sklearn
calls, operations outside the eight primitives. The grammar catches
violations their tool cannot prevent: invalid operations blocked at call
time, before they produce a number. Their tool is post-hoc; the grammar
is a guardrail. The two are complementary.

\subsection{sklearn API design}\label{sklearn-api-design}

Buitinck et al. (\citeproc{ref-buitinck2013sklearn}{2013}) formalized
the fit/predict/transform protocol that sklearn standardized as a
duck-typed typestate. The grammar extends this chain: sklearn enforces
one transition (unfitted → fitted); the grammar enforces a multi-state
chain (untagged → partitioned → fitted → assessed) covering the full
assessment lifecycle, with an explicit \texttt{NotFittedError}-style
guard at every primitive boundary.

\subsection{mlr3}\label{mlr3}

Binder et al. (\citeproc{ref-binder2021mlr3pipelines}{2021}) is the most
formally specified existing framework: typed PipeOp inputs and outputs
with graph-based composition. However, mlr3's types serve composition
(how to build pipelines) rather than validity (what workflows are
rejected based on epistemic grounds). mlr3 defines no rejection
criterion; no workflow pattern is structurally invalid. The two are
complementary: mlr3's engineering could serve as an implementation
substrate for the grammar's constraints.

\subsection{The novelty claim}\label{the-novelty-claim}

The individual primitives (split, fit, evaluate, predict) are present in
every ML framework. Their enumeration is not novel. The grammar
contributes three things that, in combination, are:

\begin{enumerate}
\def\labelenumi{\arabic{enumi}.}
\item
  \textbf{The terminal guard.} To my knowledge, no other ML framework
  documented in the peer-reviewed methodology literature enforces the
  evaluate/assess boundary at call time (as of May 2026). Repeated
  test-set assessment is a runtime failure in a conforming
  implementation, not a style violation. \(K\)-metering on the
  validation partition provides a soft diagnostic of accumulated
  optimization pressure for the practitioner and for audit (see
  §Optimization Leakage for the regime applicability).
\item
  \textbf{A specification.} The Appendix defines the type system,
  guards, and validity conditions precisely enough for independent
  reimplementation. To my knowledge, no existing supervised-learning
  framework publishes a conformance-checkable specification of
  correctness conditions.
\item
  \textbf{Empirical grounding.} Each constraint is connected to a
  measured effect size (\citeproc{ref-roth2026landscape}{Roth 2026}):
  peeking at \(K = 10\) produces \(d_z\) = 0.929, seed inflation follows
  \(\log(K)\) (\(R^2 > 0.99\)), and at practical sizes the noise is
  \(\sim\!90\%\) of the measured effect. The constraints target what
  costs the most, not what sounds worst.
\end{enumerate}

The available defences span complementary layers: documentation,
lint-style detection (\citeproc{ref-yang2022leakage}{Yang et al. 2022};
\citeproc{ref-alomar2025leakagedetector}{AlOmar et al. 2025}), formal
static analysis (\citeproc{ref-drobnjakovic2024abstract}{Drobnjaković,
Subotić, and Urban 2024}), call-time guards (this grammar), and
dependent types. Each trades rejection time against coverage scope. The
grammar's runtime-guard position trades coverage (only the eight
primitives) for earlier rejection (before the leaky number exists);
static analysis trades later detection for broader code coverage. The
grammar is the strongest runtime-guard enforcement available in
dynamically typed languages without external tooling.

\section{Empirical Motivation}\label{sec-empirical}

The grammar's formal properties are verifiable from the specification
alone; a reader can check the Appendix without running a single
experiment. But a grammar that rejects workflows nobody cares about is
useless. A companion study (\citeproc{ref-roth2026landscape}{Roth 2026})
across 2,047 datasets measures how much each leakage type actually
costs, so the constraints can be judged by both logical necessity and
empirical severity:

\begin{longtable}[]{@{}
  >{\centering\arraybackslash}p{(\linewidth - 6\tabcolsep) * \real{0.0700}}
  >{\raggedright\arraybackslash}p{(\linewidth - 6\tabcolsep) * \real{0.2500}}
  >{\raggedright\arraybackslash}p{(\linewidth - 6\tabcolsep) * \real{0.3000}}
  >{\raggedright\arraybackslash}p{(\linewidth - 6\tabcolsep) * \real{0.3800}}@{}}
\toprule\noalign{}
\begin{minipage}[b]{\linewidth}\centering
Class
\end{minipage} & \begin{minipage}[b]{\linewidth}\raggedright
Mechanism
\end{minipage} & \begin{minipage}[b]{\linewidth}\raggedright
Measured effect
\end{minipage} & \begin{minipage}[b]{\linewidth}\raggedright
Grammar response
\end{minipage} \\
\midrule\noalign{}
\endhead
\bottomrule\noalign{}
\endlastfoot
I & Estimation (normalize before split) & \(d_z \approx 0\);
\(|\Delta\)AUC\(| < 0.005\) & Constraint 2 (principled, costless) \\
II & Selection (peek, seed, screen) & \(d_z = 0.27\)--\(0.93\);
\(\Delta\)AUC \(= +0.013\)--\(0.045\) & Constraints 1, 3, 4 + terminal
assess \\
III & Memorization (train on eval data) & \(d_z = 0.29\)--\(1.38\);
\(\Delta\)AUC \(= +0.001\)--\(0.073\) & Constraint 4 (provenance
gate) \\
IV & Boundary (partition \(\neq\) deployment) & \(d_z\)
domain-dependent; \(\Delta\)AUC \(\approx +0.023\) where drift exists &
\texttt{split\_temporal}, \texttt{split\_group} \\
\end{longtable}

The four classes are defined by mechanism
(\citeproc{ref-roth2026landscape}{Roth 2026}); the grammar targets the
classes with large effects. Class I effects are negligible at any sample
size, consistent with Bousquet and Elisseeff
(\citeproc{ref-bousquet2002stability}{2002})'s stability theory. Every
selection mechanism decomposes into noise exploitation (decaying as
\(1/\sqrt{n}\)) and genuine diversity (the true performance spread
across the selection pool). At the corpus median
(\(n \approx 1{,}900\)), the measured Class II effect is consistent with
approximately 90\% noise exploitation under the seed-as-zero-diversity
modelling assumption (Roth (\citeproc{ref-roth2026landscape}{2026}),
Limitation 11), reflecting selection bias that inflates reported scores.
Under any violation of the i.i.d. assumption (temporal shift, group
structure, spatial correlation), the inner-CV optimism becomes
structural bias that does not decay with \(n\), because more data from a
biased sampling process does not correct the bias. Random
cross-validation does not prevent this structural leakage; it censors
the evidence by destroying the structure that would reveal it. The
grammar's boundary-aware primitives (\texttt{split\_temporal},
\texttt{split\_group}, \texttt{cv\_temporal}, \texttt{cv\_group})
address the structural layer; the terminal assess gate addresses the
selection layer on top. The measured leakage types are samples from an
infinite category of provenance violations \emph{within the grammar's
scope}: the grammar prevents the category, not just the measured samples
(this follows by construction from the guard definitions, not from
empirical coverage). Contamination baked into the raw data before any
primitive touches it (a feature that encodes the target, a generating
process that shifted silently) is invisible to the provenance registry.
Class III effects are capacity-dependent: decision trees memorize more
than random forests. Full effect sizes, decomposition analysis, and
confidence intervals are reported in the companion study
(\citeproc{ref-roth2026landscape}{Roth 2026}).

The design is within-subject counterfactual: a clean workflow and a
leaky workflow run on the same data, the same folds, the same seed. Only
the leakage perturbation differs. Both outcomes are observed for every
dataset, so each \(\Delta\)AUC is an individual treatment effect. All
AUC values are 5-fold stratified CV AUCs from out-of-fold predictions;
the leakage perturbations contaminate the training process (pre-split
scaling, duplicate injection, peek-driven CV-variant selection), not the
prediction step itself, so OOF AUC is inflated even though predictions
remain formally out-of-fold. Moreover, the 2,047 datasets were split at
the \emph{dataset level} into discovery (\(k = 1{,}007\)) and
confirmation (\(k = 1{,}040\)) halves by deterministic MD5 hash, fixed
before analysis. Every finding replicates on the independent
confirmation half with zero failures. Three grammar predictions were
tested before observing results: two confirmed, one falsified
(\citeproc{ref-roth2026landscape}{Roth 2026}).

\newpage

\section{Discussion}\label{discussion}

\subsection{What the grammar does not
do}\label{what-the-grammar-does-not-do}

A grammar that prevents everything prevents nothing useful: it would
reject valid workflows alongside invalid ones. The grammar's scope is
deliberately narrow: structural errors where the evaluation pipeline is
compromised. Semantic errors (choosing a wrong algorithm, a misleading
metric, or an inappropriate model for the domain) are the practitioner's
judgment, not the grammar's jurisdiction:

\begin{longtable}[]{@{}
  >{\raggedright\arraybackslash}p{(\linewidth - 2\tabcolsep) * \real{0.5500}}
  >{\raggedright\arraybackslash}p{(\linewidth - 2\tabcolsep) * \real{0.4500}}@{}}
\toprule\noalign{}
\begin{minipage}[b]{\linewidth}\raggedright
Valid in the grammar but poor in practice
\end{minipage} & \begin{minipage}[b]{\linewidth}\raggedright
Why the grammar allows it
\end{minipage} \\
\midrule\noalign{}
\endhead
\bottomrule\noalign{}
\endlastfoot
Logistic regression on 1M rows when XGBoost dominates & Algorithm choice
is not structural \\
Accuracy on a 99/1 imbalanced dataset & Metric selection is not
structural \\
\(k\)-fold CV on time-series data & Temporal awareness requires domain
knowledge\footnote{Configuring boundary-aware CV involves methodological
  trade-offs (block size, environmental gradients); Roberts et al.
  (\citeproc{ref-roberts2017cross}{2017}) survey these for spatial,
  temporal, hierarchical, and phylogenetic dependence structures and
  document that aggressive blocking can over-estimate interpolation
  error.} \\
Normalizing binary features & Feature-level decisions are semantic \\
\end{longtable}

This is the ML analogue of Chomsky's distinction between syntactic and
semantic well-formedness. ``Colorless green ideas sleep furiously'' is
grammatically valid but semantically nonsense
(\citeproc{ref-chomsky1957syntactic}{Chomsky 1957}). A workflow that
passes all type checks but produces a poor model for semantic reasons is
the ML analogue. Grammatical compliance is necessary for structural
validity but not sufficient for scientific quality. A diagnostics layer
(\texttt{check}, \texttt{drift}, \texttt{profile}, \texttt{enough})
addresses semantic quality as a separate concern.

Structural validity does not imply numerical reproducibility.
\texttt{fit(data,\ seed=42)} and \texttt{fit(data,\ seed=43)} are
grammatically identical; they may produce models with different metrics
and different conclusions. The grammar prevents structural leakage at
the language-API boundary; reproducibility is a structural correctness
concern at the infrastructure layer (seeds, environment versioning,
hardware determinism). Both are structural at different layers, and they
compose: a workflow that is grammar-valid and
reproducibility-disciplined is structurally correct end-to-end. The
grammar does not subsume the infrastructure layer, only what is
enforceable at the eight-primitive boundary.

The primitive set is operation-centric: all eight primitives are verbs.
This is a design choice, not an oversight. Validity constraints are
enforced at operations, not at values: a DataFrame is neither valid nor
invalid in isolation; it becomes invalid when passed to \texttt{fit}
without a partition tag. The types (DataFrame, Partition, Model,
Metrics) are defined formally in the Appendix and are co-equal in the
specification; they appear secondary in the prose because the actionable
grammar lives on the verbs, not the values flowing between them.

\subsection{Small N}\label{small-n}

Below approximately \(N = 30\), the grammar's three-way split becomes
statistically vacuous. The split does not disappear; it contracts: the
test partition is reabsorbed into the leave-one-out rotation, where
every observation serves as the test exactly once. \texttt{split}
returns a Partition with \texttt{test} = \(\emptyset\) and
\texttt{assess} returns the LOO-aggregated estimate; the practitioner
has no separate terminal holdout. At \(N < 10\) the design has no
statistical justification at all and the grammar's preconditions are
effectively vacuous.

\begin{longtable}[]{@{}
  >{\raggedright\arraybackslash}p{(\linewidth - 4\tabcolsep) * \real{0.1500}}
  >{\raggedright\arraybackslash}p{(\linewidth - 4\tabcolsep) * \real{0.3000}}
  >{\raggedright\arraybackslash}p{(\linewidth - 4\tabcolsep) * \real{0.5500}}@{}}
\toprule\noalign{}
\begin{minipage}[b]{\linewidth}\raggedright
\(N\)
\end{minipage} & \begin{minipage}[b]{\linewidth}\raggedright
Grammar mode
\end{minipage} & \begin{minipage}[b]{\linewidth}\raggedright
Recommendation
\end{minipage} \\
\midrule\noalign{}
\endhead
\bottomrule\noalign{}
\endlastfoot
\(< 10\) & Does not apply & Exact tests, Bayesian, case studies \\
\(10\)--\(29\) & LOO-CV, one model & Report massive uncertainty \\
\(30\)--\(199\) & \(k\)-fold CV, 1--3 models & No screening, no
stacking \\
\(200\)--\(999\) & Full grammar feasible & Screen, tune, holdout \\
\(\geq 1000\) & Grammar in its element & All strategies available \\
\end{longtable}

The deeper constraint is not raw \(N\) but the ratio of events to
predictors: logistic regression needs approximately 10--20 events per
predictor, codified as a textbook rule by Harrell
(\citeproc{ref-harrell2015regression}{2015}); modern refinements for
continuous and binary outcomes are in Riley et al.
(\citeproc{ref-riley2019minimum}{2019a}) and Riley et al.
(\citeproc{ref-riley2019minimumpart2}{2019b}), while modern modelling
techniques (SVM, neural nets, random forests) ``may need over 10 times
as many events per variable'' (\citeproc{ref-vanderploeg2014modern}{van
der Ploeg, Austin, and Steyerberg 2014}), and still show optimism even
above 200 events per variable. The grammar's real contribution at small
\(N\) is discipline: separate what you learn on from what you judge on.
That discipline applies at every \(N\). The mechanism of separation
requires a minimum investment of observations.

\subsection{External validation}\label{sec-external-validation}

External validation is the missing piece and the paper's most important
open question. This paper is a specification paper with reference
implementations: the type system, guards, and conformance conditions are
verifiable from the specification alone. Whether the grammar reduces
leakage errors in practice is a separate empirical question that the
specification does not answer. Wickham had ggplot2 users before the
grammar paper; here the sequence is reversed. A between-subjects
experiment comparing leakage rates in student code written with the
grammar versus sklearn, scored by an automated checker blind to
condition, would test the central claim directly. The relevant effect
size is not the leakage magnitude (\(d_z\) = 0.929 for Class II) but the
difference in leakage \emph{rates} between conditions, a behavioral
quantity that is unknown until the study is run. If the grammar
eliminates leakage in the treatment arm entirely (the structural
guarantee predicts this), the effect size depends on the baseline
leakage rate in the control arm. At a conservative \(d = 0.8\), a
two-group design with \(n = 25\) per arm achieves 80\% power at
\(\alpha = 0.05\); the study is feasible at classroom scale. If the
grammar does not measurably reduce leakage in human-written code, the
type system is correct but practically inert. That outcome is possible
and would be informative.

\subsection{Enforcement gaps}\label{sec-enforcement-gaps}

All implementations use gates, not funnels: \texttt{fit},
\texttt{evaluate}, and \texttt{assess} reject data that has not passed
through \texttt{split}. The tradeoff: \texttt{ml.fit(df,\ "y")} without
a prior \texttt{split} call no longer works by default;
\texttt{config(guards="off")} provides an explicit, auditable exit
analogous to Rust's \texttt{unsafe} blocks.

The provenance system uses \emph{content-addressed partition identity}:
\texttt{split} computes a deterministic fingerprint of each output
partition and registers it in a session-scoped provenance registry.
Guards query the registry by content fingerprint, not metadata
attribute, so provenance survives host-language operations that strip
attributes (\texttt{merge}, \texttt{concat}, \texttt{groupby} in pandas;
\texttt{dplyr}, \texttt{rbind} in R). This closed the most significant
structural gap in prior versions (\emph{tag erasure}), where typed
metadata was silently discarded by untyped operations in the surrounding
environment.

Five known gaps remain. (1) The registry is session-scoped; a kernel
restart resets the assess-once counter. This is intentional: the
grammar's guarantee is \emph{within a research session}, sufficient for
implementation errors and invalid orderings within a workflow.
Cross-session settings (LLM agents per prompt, AutoML retraining
schedules, multi-team pipelines) require a broader workflow grammar that
persists provenance across sessions; that outer layer is a separate
paper. (2) DataFrames with unhashable column types bypass fingerprinting
silently, the one case where the guard architecture has a genuine blind
spot rather than a conscious tradeoff. (3) The assess-once constraint is
per holdout, not per model: a practitioner who trains 5 models and
assesses each on the same test set has performed model selection on test
data. The design accepts this tradeoff: model selection belongs in the
evaluate zone, and the registry marks the holdout as spent after the
first assessment regardless of model identity. (4) \emph{Re-split
bypass.} The registry tracks partition identity by content fingerprint,
not split ancestry, so
\texttt{split(df,\ seed=42);\ assess;\ split(df,\ seed=99);\ assess}
passes both guards (each partition has a fresh fingerprint).
Within-session assess-once prevents composition leakage from internal
tuning, selection, and peeking; across sessions, Dwork's
adaptive-data-analysis framework (\citeproc{ref-dwork2015reusable}{Dwork
et al. 2015}) budgets multi-query reuse, complementary at a different
scale (orchestrator layer above). Operators should be aware that
re-splitting the same raw data is structurally invisible to the
within-session guard. (5) \emph{Feature derivation outside the typed
primitives.} The grammar rejects
\texttt{select\_features(X,\ y)\ →\ split\ →\ fit} (the unregistered
DataFrame from \texttt{select\_features} cannot enter \texttt{fit}), but
manual feature derivation \emph{before} \texttt{split} (e.g.,
\texttt{df{[}\textquotesingle{}feat\textquotesingle{}{]}\ =\ derived\_from\_target(df);\ split(df)})
is silently allowed: from the grammar's perspective the input to
\texttt{split} is just a DataFrame, and the leaky column is invisible.
The guard architecture operates on the eight primitives' boundary, not
on the contents of the DataFrames they receive.

A grammar is a specification; any implementation is an approximation of
it. Early SQL implementations violated Codd's relational model
routinely; the specification remained correct, and implementations
improved over decades.

\textbf{Committed falsifying workflow.} Given the gaps above, the
in-scope structural guarantee is: \emph{a workflow expressed exclusively
through the eight primitives, in a single Python or R session, on a
content-addressable DataFrame whose \texttt{split} is called once, with
no manual feature derivation between source-DataFrame load and
\texttt{split}, will not produce an Evidence value carrying Class I--IV
leakage (structural leakage through guards).} Optimization leakage from
evaluate-set thrashing (selection pressure accumulated on
\texttt{s.valid} before a single \texttt{assess} call) is a separate
regime addressed by \(K\)-metering as a soft diagnostic (see
§Optimization Leakage), not by structural rejection. The structural
guarantee is falsified by the demonstration of any such workflow that
nonetheless produces Class I--IV leakage detectable by the landscape
paper's experimental procedure on a held-out oracle dataset. Workflows
outside this scope (re-split, manual feature mutation, multi-session,
kernel restart, unhashable columns, model selection across multiple
\texttt{assess} calls on different partitions of the same data) are
documented gaps.

\subsection{Production and deployment}\label{production-and-deployment}

Production retraining follows the same \texttt{split} → \texttt{cv} →
\texttt{fit} → \texttt{evaluate} → \texttt{assess} workflow; existing
guards apply without modification. Under concept drift,
\texttt{Evidence} measures generalization to the training distribution,
not the current environment; the reference implementations include a
\texttt{drift} diagnostic verb for monitoring distributional shift, kept
outside the eight primitives by design.

\subsection{Optimization leakage}\label{sec-optimization-leakage}

The grammar closes data leakage but does not fully close optimization
leakage: a practitioner who runs \texttt{screen} with 11 algorithms,
\texttt{tune} with 500 configurations, and then calls \texttt{assess}
has made hundreds of model selection decisions on the validation set.
The expected maximum of \(K\) noisy evaluations grows with the number of
decisions made on the validation set; Cawley and Talbot
(\citeproc{ref-cawley2010overfitting}{2010}) analyzed the
model-selection overfitting this reflects.

The grammar addresses this through progressive access control. The three
partitions operate at different speeds:

\begin{enumerate}
\def\labelenumi{\arabic{enumi}.}
\item
  \textbf{Train (unrestricted).} Iteration within the training partition
  is unlimited and honest: cross-validation folds are ephemeral, created
  and consumed within the inner loop.
\item
  \textbf{Valid (metered).} Each \texttt{evaluate} call increments a
  per-partition counter \(K\) in the provenance registry. The closest
  prior art is Blum and Hardt (\citeproc{ref-blum2015ladder}{2015}), who
  metered holdout access in ML competitions; the present mechanism
  applies the same principle inside the practitioner's workflow.
  \(K\)-metering is a soft diagnostic, not a hard gate: no call is
  rejected by \(K\). The count reflects real inflation pressure in the
  weakly-correlated regime (seed cherry-picking, hyperparameter random
  search across heterogeneous configs), where the expected inflation
  grows with \(\log K\). In the strongly-correlated regime (e.g.,
  screening 11 algorithms on the same CV folds, landscape's screen
  experiment) the count overstates pressure because correlated maxima
  saturate immediately and effective \(K\) stays near 1. In both regimes
  the load-bearing prevention is the structural guards on the test
  partition; \(K\)-metering makes evaluation pressure visible to the
  practitioner and to audit.
\item
  \textbf{Test (terminal).} One assessment per holdout. Enforced by the
  assess-once constraint.
\end{enumerate}

This progressive slowdown (free on train, metered on valid, terminal on
test) mirrors the natural structure of disciplined research: explore
freely, steer occasionally, commit once. The assess-once constraint is
the zero-reuse limit of Dwork et al.
(\citeproc{ref-dwork2015reusable}{2015})'s adaptive framework
(\(\epsilon = 0\)): rather than permitting controlled repeated access
with formal privacy guarantees, it permits no reuse at all. This trades
the reusable holdout's flexibility for enforceability in a static type
system: engineering discipline rather than a privacy guarantee.

\newpage

\section{Conclusion}\label{conclusion}

The evaluate/assess boundary is a constraint this grammar enforces, not
a textbook convention left to discipline. Its type system remembers the
conventions so you don't have to. Time currently spent on unsafe
exploration (fitting on test data, re-assessing holdouts until results
look favorable, cherry-picking seeds and screens) is forced into the
evaluate zone, where iteration is unlimited but metered, and therefore
honest. The grammar also rejects pre-split preprocessing for structural
completeness, though the companion study finds this leakage class is
empirically negligible at typical iid tabular dataset sizes: a textbook
warning that holds in form but rarely costs much in practice. The
grammar treats it as a structural completeness item rather than a
load-bearing prevention; the load-bearing prevention is at the
evaluate/assess boundary.

Eight primitives, four constraints, two reference implementations
(Python, R), and a companion study across 2,047 datasets
(\citeproc{ref-roth2026landscape}{Roth 2026}): the grammar is internally
consistent and the implementations satisfy the conformance conditions.
At practical dataset sizes, the leakage classes the grammar rejects
produce substantial effects, consistent with an approximately 90\%
noise-exploitation share (under the seed-as-zero-diversity decomposition
assumption flagged in (\citeproc{ref-roth2026landscape}{Roth 2026},
Limitation 11)) that is empirically grounded and preventable at call
time by rejection. Under any violation of the i.i.d. assumption (a
common regime in real deployments), the selection bias becomes
irreducible because past evaluation is a biased estimator of future
performance, and random cross-validation censors the evidence by
destroying the structure that would reveal it. The grammar addresses
both layers: boundary-aware splitting for the structural contamination,
and terminal assess as the final safety net for assumption violations
and implementation bugs that no earlier layer caught. Three directional
predictions recorded before observing results: two confirmed, one
falsified. The grammar is falsifiable and has survived its initial
tests.

Whether the grammar reduces leakage errors in practice remains the
central empirical question (Section~\ref{sec-external-validation}). As
pipelines are increasingly generated rather than written \emph{within a
single execution context}, the grammar's role shifts from constraining
the human to constraining the generator at that horizon; extending the
same guarantee across sessions (AutoML schedules, LLM agents called
fresh per prompt) requires the outer workflow grammar deferred in
Section~\ref{sec-enforcement-gaps}.

Leakage and fairness bias are distinct failures with distinct root
causes (fairness bias surveyed in (\citeproc{ref-roth2022biased}{Roth
2022})), but a model whose training process is contaminated by test data
cannot be trusted to measure fairness either; the contaminated
measurement is what any downstream fairness analysis builds on. The
grammar does not detect fairness bias, but it closes a methodological
precondition for honest fairness measurement: honest measurement
requires honest evaluation, and honest assessment is what the type
system enforces.

A researcher who reports \texttt{GridSearchCV.best\_score\_} as their
model's accuracy is not p-hacking; the held-out test is one method call
further than the default API output. A social scientist who selects
predictors on the pooled sample before cross-validating is not
unfamiliar with overfitting; their statistics package presented variable
selection and validation as independent steps. An economist who reports
in-sample \(R^2\) is not ignorant of holdout methodology; the holdout
was two menu options deeper buried than the default workflow. An
automated retraining pipeline that refits monthly on all available data,
including the month it will immediately score, is not malfunctioning; it
is fulfilling its update schedule. A language model asked to generate a
pipeline will select the best out of 100 model families on the test set,
because that is the modal pattern in its training distribution, and no
gradient pointed away from it. The 648 papers with leakage errors are
not anomalies; they are the expected output of a field whose
methodological knowledge outpaced its tooling.

The grammar does not require the field to learn something new. It
requires the tools to enforce what the field already knows. The training
boundary should be a first-class type, and leakage should be a
violation, not a matter of professional discipline. Professional
discipline varies across career stages, institutions, and time zones.
Type systems do not. Within a single execution context (one notebook,
one CI run, one prompt-to-pipeline LLM call), structural enforcement
closes the gap that professional discipline alone cannot. Extending that
guarantee to multi-session settings requires the cross-session workflow
grammar discussed in Section~\ref{sec-enforcement-gaps} and left to
future work. Investment allocation, credit scoring, medical diagnosis,
recidivism prediction: when the pipeline that produces the number runs
faster than any methodology review can follow, knowing the rules is not
enough; the tool must enforce them at call time.

\newpage

\section*{Appendix: Formal
Specification}\label{appendix-formal-specification}
\addcontentsline{toc}{section}{Appendix: Formal Specification}

\subsection{Type system}\label{type-system}

The grammar is a typed specification over nine types. Each primitive has
a type signature, a guard (precondition checked at call time), and an
effect (state mutation after call). Three types are terminal (their
values feed no further primitive):

\begin{longtable}[]{@{}
  >{\raggedright\arraybackslash}p{(\linewidth - 4\tabcolsep) * \real{0.1500}}
  >{\raggedright\arraybackslash}p{(\linewidth - 4\tabcolsep) * \real{0.7500}}
  >{\centering\arraybackslash}p{(\linewidth - 4\tabcolsep) * \real{0.1000}}@{}}
\toprule\noalign{}
\begin{minipage}[b]{\linewidth}\raggedright
Type
\end{minipage} & \begin{minipage}[b]{\linewidth}\raggedright
Structure
\end{minipage} & \begin{minipage}[b]{\linewidth}\centering
Terminal
\end{minipage} \\
\midrule\noalign{}
\endhead
\bottomrule\noalign{}
\endlastfoot
DataFrame & Tabular data with named columns and
\texttt{partition\_tag:\ \{None,\ train,\ valid,\ test,\ dev\}}. Freshly
loaded data carries tag \texttt{None} (untagged). \texttt{split} assigns
tags \texttt{train}, \texttt{valid}, \texttt{test} to the output
partitions. The \texttt{dev} tag is assigned to the union of train and
valid (\texttt{s.dev}). & No \\
Partition & \texttt{\{train,\ valid,\ test,\ dev:\ DataFrame\}} where
\texttt{dev} = train \(\cup\) valid & No \\
CVResult &
\texttt{\{folds:\ {[}(train\_idx,\ valid\_idx){]}\^{}k,\ k:\ N,\ target:\ str\}}.
Produced by \texttt{cv} from a training or dev DataFrame; blocks
\texttt{.train}, \texttt{.valid}, \texttt{.test}, \texttt{.dev} access.
Test partition stays on the originating Partition. & No \\
PreparedData &
\texttt{\{data:\ DataFrame\_numeric,\ state:\ Transformer,\ target:\ str,\ task:\ \{clf,\ reg\}\}}
& No \\
Model &
\texttt{\{algorithm:\ str,\ task:\ \{clf,\ reg\},\ fitted:\ bool\}}\textsuperscript{‡}
& No \\
Predictions & One-column numeric DataFrame & Yes\textsuperscript{†} \\
Metrics & \texttt{str\ →\ float} & Yes \\
Evidence & Sealed named type wrapping \texttt{str\ →\ float}; must be a
class distinct from Metrics in the host language, not implicitly
substitutable for Metrics under structural or duck typing. No primitive
accepts Evidence as input. & Yes \\
Explanation & \texttt{str\ →\ float} (importances or partial
dependences) & Yes \\
\end{longtable}

\textsuperscript{†} Predictions is not terminal in stacking strategies:
the \texttt{stack} strategy reshapes out-of-fold predictions into a new
DataFrame and passes it to a second \texttt{fit} call (the
meta-learner). In the grammar's 8-primitive DAG, no primitive accepts
Predictions directly as a typed input; the reshape is performed by the
strategy orchestration, not a primitive.

\textsuperscript{‡} The implementations additionally track
\texttt{assessed} as an integer counter on the model
(\texttt{\_assess\_count} in Python, \texttt{assess\_count} in R) for
fast-path diagnostics. The structurally load-bearing enforcement is
per-holdout: the provenance registry tracks which test partitions have
been assessed, and the guard rejects any subsequent call on the same
partition regardless of model identity.

Two embedded types appear in the type table above but are not primitive
types:

\begin{itemize}
\tightlist
\item
  \textbf{\texttt{DataFrame\_numeric}}: a \texttt{DataFrame} where every
  column is numeric; a structural subtype of \texttt{DataFrame} (all
  rows of \texttt{DataFrame\_numeric} are valid \texttt{DataFrame}
  values but not vice versa). Produced by \texttt{prepare}; consumed by
  model algorithms.
\item
  \textbf{\texttt{Transformer}}: an interface type satisfying
  \texttt{state.transform(X:\ DataFrame)\ →\ DataFrame\_numeric}. It
  encodes the fit-time preprocessing state so that validation data can
  be transformed consistently without re-fitting. Not a free-standing
  type; it is a field type embedded in \texttt{PreparedData}.
\item
  \textbf{\texttt{partition\_tag}}: an enum attribute
  \texttt{\{None,\ train,\ valid,\ test,\ dev\}} on every DataFrame. Not
  a separate type; it is an attribute of the \texttt{DataFrame} type.
  Freshly loaded data carries \texttt{None}; \texttt{split} assigns
  \texttt{train}, \texttt{valid}, \texttt{test}; accessing
  \texttt{s.dev} (train \(\cup\) valid) assigns \texttt{dev}.
\end{itemize}

\subsection{Primitive operations}\label{primitive-operations}

Each primitive \(\sigma\) has a type signature, a guard \(G(\sigma)\)
(precondition checked at call time), and an effect \(E(\sigma)\) (state
mutation after call). \textbf{Registered} means the DataFrame's content
fingerprint appears in the session-scoped provenance registry with a
valid \texttt{partition\_tag} assigned by a prior \texttt{split} call:

\begin{longtable}[]{@{}
  >{\raggedright\arraybackslash}p{(\linewidth - 8\tabcolsep) * \real{0.1200}}
  >{\raggedright\arraybackslash}p{(\linewidth - 8\tabcolsep) * \real{0.1500}}
  >{\raggedright\arraybackslash}p{(\linewidth - 8\tabcolsep) * \real{0.1300}}
  >{\raggedright\arraybackslash}p{(\linewidth - 8\tabcolsep) * \real{0.3500}}
  >{\raggedright\arraybackslash}p{(\linewidth - 8\tabcolsep) * \real{0.2500}}@{}}
\toprule\noalign{}
\begin{minipage}[b]{\linewidth}\raggedright
Primitive
\end{minipage} & \begin{minipage}[b]{\linewidth}\raggedright
Input
\end{minipage} & \begin{minipage}[b]{\linewidth}\raggedright
Output
\end{minipage} & \begin{minipage}[b]{\linewidth}\raggedright
Guard \(G\)
\end{minipage} & \begin{minipage}[b]{\linewidth}\raggedright
Effect \(E\)
\end{minipage} \\
\midrule\noalign{}
\endhead
\bottomrule\noalign{}
\endlastfoot
\textbf{split} & DataFrame, params & Partition & --- & assigns
\texttt{partition\_tag\ in\ \{train,\ valid,\ test,\ dev\}} to output
partitions; registers content fingerprints of all partitions (including
\texttt{dev} = train \(\cup\) valid) in the provenance registry with
their tags and split lineage \\
\textbf{cv} & DataFrame, params & CVResult & \texttt{data} registered by
\texttt{split}; \texttt{data.partition\_tag\ in\ \{train,\ dev\}} &
blocks \texttt{.train}, \texttt{.valid}, \texttt{.test}, \texttt{.dev}
on result; test stays on originating Partition \\
\textbf{prepare} & DataFrame, target & PreparedData & \texttt{data}
registered by \texttt{split};
\texttt{data.partition\_tag\ in\ \{train,\ valid,\ dev\}} & fits
Transformer on data; stores in \texttt{PreparedData.state} \\
\textbf{fit} & DataFrame \(\vert\) CVResult \(\vert\) PreparedData,
target & Model & \texttt{data} registered by \texttt{split}; if
DataFrame, \texttt{data.partition\_tag\ in\ \{train,\ valid,\ dev\}}; if
CVResult, produced by \texttt{cv}; if PreparedData, inherits provenance
from its source DataFrame & \texttt{model.fitted\ ←\ true}; prepare
applied per fold (skipped if input is PreparedData) \\
\textbf{predict} & Model, DataFrame & Predictions &
\texttt{model.fitted\ =\ true} & --- \\
\textbf{evaluate} & Model, DataFrame & Metrics &
\texttt{model.fitted\ =\ true}; \texttt{data} registered by
\texttt{split}; \texttt{data.partition\_tag\ !=\ test} & --- \\
\textbf{explain} & Model {[}, DataFrame{]} & Explanation &
\texttt{model.fitted\ =\ true} & --- \\
\textbf{assess} & Model, DataFrame & Evidence & \texttt{data} registered
by \texttt{split}; \texttt{data.partition\_tag\ =\ test}; test partition
not previously assessed in provenance registry & mark test partition as
assessed in registry; \texttt{model.assessed\ ←\ true} (diagnostic) \\
\end{longtable}

\subsection{When is a workflow valid?}\label{when-is-a-workflow-valid}

A workflow (a sequence of primitive calls) is \textbf{valid} if:

\begin{enumerate}
\def\labelenumi{\arabic{enumi}.}
\tightlist
\item
  \textbf{Types connect:} each primitive's output type matches the next
  primitive's expected input type.
\item
  \textbf{Guards pass:} each primitive's precondition holds at call
  time, given all prior effects.
\item
  \textbf{Effects apply in order:} state mutations (e.g., test partition
  marked as assessed in registry) accumulate sequentially.
\end{enumerate}

A workflow is \textbf{invalid} if any step fails a type check or a
guard. The four hard constraints are guards in this sense: not rules a
user must remember, but checks that reject invalid workflows at call
time.

\textbf{Note on evaluate guard.} The guard on \texttt{evaluate} rejects
test-tagged data; it accepts train and valid partitions. Evaluating on
training data is legitimate (train-vs-valid score comparison for
overfitting diagnosis). The guard prevents test data from entering the
iterate cycle: if \texttt{evaluate} accepted test data, the practitioner
could iterate on test-set feedback, which is structurally equivalent to
training on the test set. This is a design decision: \texttt{evaluate}
is the practice exam (repeatable, on validation data); \texttt{assess}
is the final exam (terminal, on test data).

\textbf{Note on implementation separation.} \texttt{assess} and
\texttt{evaluate} are independent primitives with no call relationship.
Both delegate to a shared private scoring function for prediction and
metrics computation, but neither calls the other. \texttt{assess}
enforces the test-tag requirement and the once-per-holdout constraint;
\texttt{evaluate} enforces the test-data rejection guard. Strategy verbs
(\texttt{compare}, \texttt{validate}, \texttt{screen}, \texttt{shelf},
\texttt{tune}) that need metrics on arbitrary partitions call the same
shared scoring function directly, bypassing both \texttt{evaluate}'s and
\texttt{assess}'s guards and enforcing their own partition constraints.

\textbf{Note on branching strategies.} This definition covers linear
workflows. Branching strategies (screen, tune, stack) involve parallel
or iterated applications of the same primitives; their validity follows
from applying the same type signatures and guards to each branch
independently.

\textbf{Note on domain specializations.} \texttt{split} admits domain
specializations (\texttt{split\_temporal}, \texttt{split\_group}) that
share its type signature but carry additional guards specific to a
scientific domain. This is a representational mechanism, not a primitive
extension: the guard set \(G(\text{split})\) is parameterized by domain,
but the type signature and position in the DAG are invariant. \#\#\#
Invalid workflows as guard failures

\begin{longtable}[]{@{}
  >{\raggedright\arraybackslash}p{(\linewidth - 4\tabcolsep) * \real{0.2500}}
  >{\raggedright\arraybackslash}p{(\linewidth - 4\tabcolsep) * \real{0.3000}}
  >{\raggedright\arraybackslash}p{(\linewidth - 4\tabcolsep) * \real{0.4500}}@{}}
\toprule\noalign{}
\begin{minipage}[b]{\linewidth}\raggedright
Workflow
\end{minipage} & \begin{minipage}[b]{\linewidth}\raggedright
Failing condition
\end{minipage} & \begin{minipage}[b]{\linewidth}\raggedright
Mechanism
\end{minipage} \\
\midrule\noalign{}
\endhead
\bottomrule\noalign{}
\endlastfoot
assess(\(m\)) → assess(\(m\)) & \(G(\text{assess})\): test partition not
previously assessed & First call marks test fingerprint as assessed in
registry; second call on same holdout fails \\
assess(\(m_1\), test) → assess(\(m_2\), test) & \(G(\text{assess})\):
test partition not previously assessed & First call marks test
fingerprint as assessed in registry; second call on same holdout fails
regardless of model \\
prepare(\(X_\text{all}\)) → split → fit & \(G(\text{prepare})\): data
not registered by \texttt{split} & \texttt{prepare} rejects unregistered
data (\texttt{partition\_tag\ =\ None}); failure occurs at
\texttt{prepare}, before \texttt{split} is called \\
select\_features(\(X\), \(y\)) → split → fit & \texttt{fit} requires
split provenance & Feature selection belongs inside \texttt{prepare},
per fold (§ Feature engineering); unregistered data rejected \\
evaluate without prior fit & Type continuity & No Model in scope;
\texttt{out(split)\ =\ Partition\ !=\ in(evaluate)\ =\ Model} \\
fit(\(s\).test) & \(G(\text{fit})\):
\texttt{data.partition\_tag\ not\ in\ \{train,\ valid,\ dev\}} & Test
tag is \texttt{test}, not in \texttt{\{train,\ valid,\ dev\}}; guard
fails \\
evaluate(\(m\), \(s\).test) & \(G(\text{evaluate})\):
\texttt{data.partition\_tag\ !=\ test} & Test tag triggers guard; test
data is reserved for \texttt{assess} \\
\end{longtable}

\textbf{API maturity.} The implementations are versioned at 1.x and
follow semantic versioning. The 8 kernel primitives and 4 hard
constraints are stable; the strategy-layer verbs and their signatures
may change before 2.0. The specification, not any individual
implementation, is the stable contract.

\section*{Code and Data Availability}\label{code-and-data-availability}
\addcontentsline{toc}{section}{Code and Data Availability}

The Python implementation is available as \texttt{mlw} on PyPI
(\texttt{pip\ install\ mlw}, imported as \texttt{import\ ml}). The R
implementation is available as \texttt{ml} on CRAN and on GitHub
(\texttt{epagogy/ml}). Source code for both implementations:
\href{https://github.com/epagogy/ml}{github.com/epagogy/ml}. The
specification (Appendix A) is sufficient for independent
reimplementation.

\section*{Conflict of Interest}\label{conflict-of-interest}
\addcontentsline{toc}{section}{Conflict of Interest}

The author develops and distributes the Python and R implementations
described in this paper. The grammar specification is intended to be
implementable independently of them.

\section*{Acknowledgments}\label{acknowledgments}
\addcontentsline{toc}{section}{Acknowledgments}

This work was conducted independently and received no external funding.
I thank the colleagues and peers who provided critical feedback during
this process; they will be acknowledged individually upon journal
submission, if they choose to be named. This is ongoing work; feedback
is welcome at simon@epagogy.ai.

\section*{Disclosure}\label{disclosure}
\addcontentsline{toc}{section}{Disclosure}

Large language models (Claude, Anthropic) were used as principal
writing, analysis, and implementation tools during the preparation of
this manuscript. All scientific claims, experimental designs, empirical
results, and theoretical contributions are my own. I take full
responsibility for the content.

\section*{References}\label{references}
\addcontentsline{toc}{section}{References}

\phantomsection\label{refs}

\end{document}